\documentclass[Afour,sageh,times]{sagej}
\usepackage[ruled,linesnumbered]{algorithm2e}
\usepackage{float}
\usepackage{cases}
\usepackage{amsmath}
\usepackage{amssymb}
\DeclareGraphicsExtensions{.eps,.ps,.jpg,.bmp,.png}
\usepackage{moreverb,url}

\usepackage{amsmath,amssymb,amsfonts}
\usepackage{algorithmicx}
\usepackage{algpseudocode}
\usepackage{graphicx}
\usepackage{subfigure}
\usepackage{textcomp}
\usepackage{wrapfig}
\usepackage{enumitem}
\usepackage{siunitx}
\usepackage{xcolor}
\usepackage{booktabs}
\usepackage[normalem]{ulem}
\usepackage{cancel}

\usepackage[utf8]{inputenc}
\usepackage[english]{babel}
\usepackage{amsthm}

\newtheorem{assumption}{\hspace{1em}Assumption}
\newtheorem{problem}{\hspace{1em}Problem}

\let\vec\boldvec
\let\mat\boldvec

\usepackage[colorlinks,bookmarksopen,bookmarksnumbered,citecolor=red,urlcolor=red]{hyperref}

\newcommand\BibTeX{{\rmfamily B\kern-.05em \textsc{i\kern-.025em b}\kern-.08em
T\kern-.1667em\lower.7ex\hbox{E}\kern-.125emX}}

\def\volumeyear{2024}

\definecolor{GfLi}{RGB}{0, 0, 255}  
\definecolor{mygray}{gray}{0.5}     
\definecolor{myred}{RGB}{255, 0, 0} 
\definecolor{mygreen}{RGB}{100, 200, 100}
\definecolor{orange}{RGB}{255, 100, 100}  %

\begin{document}

\runninghead{Gaofeng Li {\it et al.}}

\title{Motion planning for highly-dynamic unconditioned reflexes based on chained Signed Distance Functions}

\author{Ken Lin\affilnum{1}, Qi Ye\affilnum{1}, Tin Lun Lam\affilnum{2}, Zhibin Li\affilnum{3}, Jiming Chen\affilnum{1}, and Gaofeng Li*\affilnum{1}}

\affiliation{\affilnum{1}College of Control Science and Engineering, Zhejiang University, Hangzhou, China.\\
\affilnum{2}Robotics and Artificial Intelligence Laboratory, The Chinese University of Hongkong, Shenzhen, China.\\
\affilnum{3}Department of Computer Science, University College London, London, The United Kingdom. \\
}

\corrauth{Gaofeng Li, College of Control Science and Engineering,
Zhejiang University,
No. 38, Zheda Road,
Xihu District,
Hangzhou, China.}

\email{gaofeng.li@zju.edu.cn}

\begin{abstract}
The unconditioned reflex (e.g., protective reflex), which is the innate reaction of the organism and usually performed through the spinal cord rather than the brain, can enable organisms to escape harms from environments. Accordingly, it is essential to endow robots the highly-dynamic unconditioned reflexes to humans and/or environments, such that it can work collaboratively with humans or serve people in daily life. Although various sampling-based and optimization-based motion planning methods have been developed for robots, their limited computational speed is not suitable for rapid motion planning in highly-dynamic environments. In this paper, we propose an online, highly-dynamic motion planning algorithm to enable highly-dynamic unconditioned reflexes for manipulators. Our method is based on a chained version of Signed Distance Functions (SDFs), which can be pre-computed and stored. Our proposed algorithm is divided into two stages. In the offline stage, we create 3 groups of local SDFs to store the geometric information of the manipulator and its working environment. The SDFs in the first group are to describe the geometric information of the static environment. The SDFs in the second group are to describe the geometric information of each link. While the SDFs in the last group are to describe the reachability of each link's end of the manipulator. In the online stage, the pre-computed local SDFs are chained together according the configuration of the manipulator, to provide global geometric information about the environment. While the point clouds of the dynamic objects serve as query points to look up these local SDFs for quickly generating escape velocity. Then we propose a modified geometric Jacobian matrix and use the Jacobian-pseudo-inverse method to generate real-time reflex behaviors to avoid the static and dynamic obstacles in the environment. The benefits of our method are validated in both static and dynamic scenarios. In the static scenario, our method identifies the path solutions with lower time consumption and  shorter trajectory length compared to existing solutions. In the dynamic scenario, our method can reliably pursue the dynamic target point, avoid dynamic obstacles, and react to these obstacles within 1 ms, which surpasses the unconditioned reflex reaction time of humans.

\end{abstract}

\keywords{Signed Distance Function(SDF), Motion planning, Unconditioned Reflex, Obstacle Avoidance}

\maketitle

\section{1. Introduction}
\label{Section1_Introduction}
\begin{figure}[ht]
    \centering
    \includegraphics[width = 0.97\hsize]{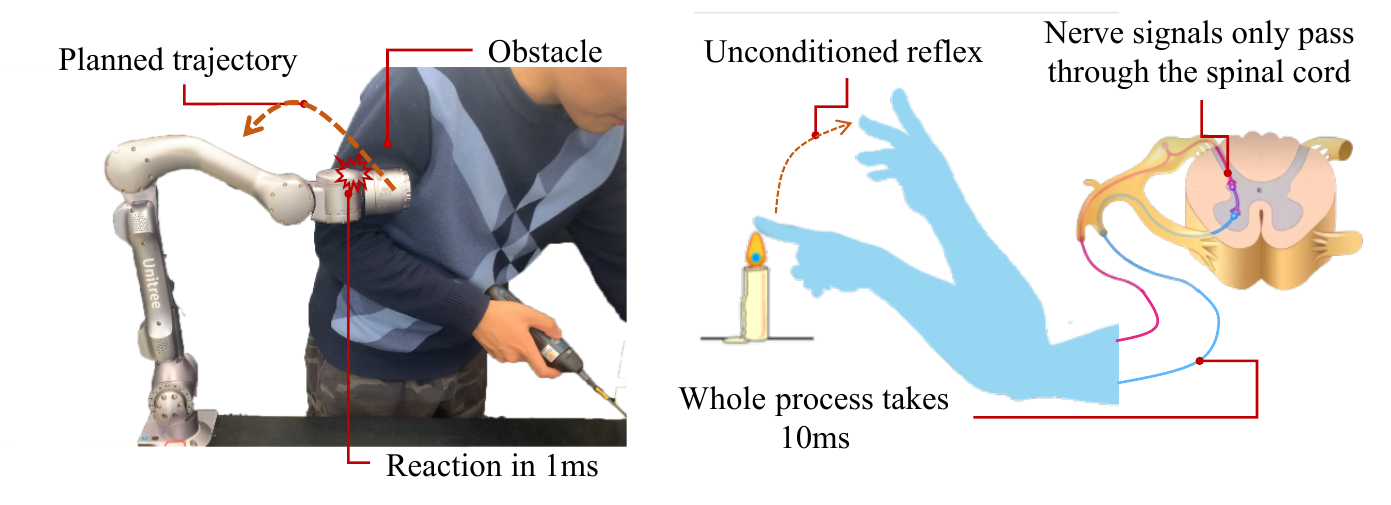}
    \caption{The unconditioned reflexes enable humans to react to stimulus (e.g., pain) quickly to avoid harm from environments. To guarantee real-time, the unconditioned reflexes are usually performed through the spinal cord rather than the brain. Accordingly, it is essential to endow robots the highly-dynamic unconditioned reflex to humans and/or environments, such that it can work collaboratively with humans or serve people in daily life. }
    \label{unconditioned_reflex}
\end{figure}

Humans exhibit both conditioned and unconditioned reflexes to respond to unexpected events in daily life. A conditioned reflex is a learned response that occurs when an irrelevant stimulus is paired with a stronger unconditioned stimulus. This process typically takes about hundreds of milliseconds, as the conditioned stimulus must be transmitted to the cerebral cortex. In contrast, an unconditioned reflex is an innate response to a specific stimulus, which involves nerve centers located below the cerebral cortex, such as in the knee-jerk reflex or hand-contraction reflex. The reflex time for unconditioned reflexes is generally around ten milliseconds (\cite{2008_International_Journal_of_Psychophysiology}, \cite{2004__International_Journal_of_Psychophysiology}). In this paper, we define highly-dynamic reflexes as the capability for fast responses and motion planning, achieving response times in the milliseconds range for both dynamic and static obstacles in the environment.

The motion planning problem is fundamental for robots (\cite{2011_IJRR_Sampling_based_algorithms}). Over past decades, many motion planning algorithms, e.g., Rapidly-exploring Random Tree (RRT, A star (A*), Artificial Potential Field (APF), have been developed to plan a collision-free path/trajectory for robots working in static environment. However, unlike traditional industrial manipulators working in structured environment, the collaborative manipulators are facing highly-dynamic environments to work collaboratively with humans or serve people in daily life. These highly-dynamic objects, like human body, bring significant challenges to the motion planning problem. Therefore, the ability to generate highly-dynamic reflexes to humans and/or environments is essential for robots to walking from structured factory to unstructured daily-life environments.

Currently, the mainstream motion planning approaches can be classified into three types: the sampling-based ones, the graph-based searching ones and the optimization-based ones, to generate a complete path for the manipulator to track (\cite{2011_IJRR_Sampling_based_algorithms}, \cite{2001_IJRR_RRT}, \cite{1996_TRO_PRM}, \cite{2020_IJRR_BIT}, \cite{2022_IJRR_AIT}).The sampling-based methods, although theoretically proven to be complete, usually require multiple iterations and batches of sampling points to generate a path. Until a complete string of sampling points connecting the start and goal is found, the manipulator cannot select the correct direction to move, which hinders the manipulator to react immediately. The graph-based searching algorithms have also been proven to be complete and consistently provide the optimal solution within the global scope. However, its searching efficiency significantly decreases in high-dimensional spaces. Additionally, similar to sampling-based methods, it requires a valid path to be determined before the manipulation can react. Similarly, the optimization-based methods have to wait for the solver to finish the optimizing procedure, which is used to generate the ideal joint trajectory based on a given objective function. Although some recent approaches had been proposed to reduce the time consumptionfor the sampling-based and optimization-based methods (\cite{2023_SR_Motion_planning_around_obstacles_with_convex_optimization}, \cite{2021_TRO_Constrained_motion_planning_networks}), these approaches inherit unavoidable drawbacks making them unsuitable for highly-dynamic environments.

In order to adapt to dynamic scenarios, the local re-planning strategy was proposed (\cite{2023_RAL_Path_Re_Planning_Design}, \cite{2013_ASCC_VF_RRT}, \cite{2018_ICRA_Collision_free_motion_planning_for_human_robot}).Based on this strategy, the manipulator moves along a preplanned trajectory. When an obstacle is detected, the local replanning strategy is invoked to avoid the obstacle using techniques such as re-sampling or the Artificial Potential Field (APF). However, these local replanning methods are essentially still sampling-based, which leads to inherent drawbacks.For example, the manipulator is unable to react to the obstacle until the whole local re-planning process is finished. In addition, the obstacle avoidance process can also take significant time for replanning, making them difficult to be applied in highly-dynamic scenarios.

The primary factor that stops the conventional motion planning algorithms being extended to dynamic scenarios is the time-consuming calculation of the distance between the dynamic obstacles and the surface of the manipulator. In conventional methods, the geometry of the obstacles and/or the manipulator are explicitly stored as a set of points/lines/triangles. (\cite{2012_ICRA_depth_space_approach_to_human_robot_collision_avoidance}, \cite{2011_RSJ_Active_camera_control_with_obstacle_avoidance}, \cite{2005_ISNN_Obstacle_Avoidance_for_Kinematically_Redundant}, \cite{2004_TOS_Obstacle_avoidance_dual_neural_network}, \cite{2006_ACC_Optimal_manipulator_path_planning}) However, the explicit representation method has no prior information about the relative localization between two different objects. Therefore, to achieve obstacle avoidance, the computation of the nearest distance between any two objects is necessary, but has to depend on a time-consuming searching for complex and irregular curves/surfaces/objects. Many approaches have been proposed to skeleton the link of the manipulator as straight lines and/or approximate the obstacles as a combination of several regular geometric shapes, to simply the calculation. However, the innate disadvantages of explicit methods make them unsuitable for highly-dynamic environments.

To overcome the time-consuming issue of explicit methods, the Signed Distance Function (SDF) (\cite{2011_ACM_Kinectfusion}, \cite{2013_RSS_Real_time_camera_tracking}, \cite{2013_RSJ_SDF_tracker}), which is an implicit method, is introduced into the motion planning problem. The SDF, which is originally proposed in computer graphics, is defined as a level set function over the whole space (\cite{1995_TPAMI_Shape_modeling_with_front_propagation}). It returns positive values on the exterior, negative values on the interior, and zero on the boundary, for a given object. In another word, the geometry of the object is embedded in the SDF values instead of being stored explicitly. Compared with explicit methods, the SDF method provides descriptions about the relative location information to the geometry for all points, including both the exteriors and interiors. Therefore, it is possible to replace the time-consuming distance calculation by a highly-efficient querying on the pre-computed SDF table at runtime (\cite{1996_ACM_volumetric_method_for_building_complex_models}).In addition, the SDF is independent of the geometric shape of the object and does not increase the query time due to the complex and non-convex shape of the object. Therefore, the SDF can be used to describe objects with any shapes. Furthermore, although it is tedious to pre-compute the SDF, the SDF only needs to be computed once for a specific geometry. These advantages make it extremely suitable for motion planning in complex scenarios. For example, \cite{2024_Arxiv_Implicit_Swept_Volume_SDF} utilized the swept volume generated by the UAV’s pre-flight trajectory as the target object to create an implicit signed distance function. They optimized the obstacle distances relative to the surface of the UAV's swept volume, enabling collision-free trajectory generation and second-level UAV path and attitude planning in complex scenarios.

However, there is still a big gap to apply the SDF in highly-dynamic scenarios. For a rigid body or a complex but static environment, it is possible to pre-computed a SDF to describe its geometric information. However, the pre-computation for articulated objects, e.g., manipulators, is challenging. The transition to articulated models results in a continuous space of signed distanced functions, as a function of the state. This means that the global SDF can no longer be strictly pre-computed. To apply the SDF, the global SDF has to be updated every time the pose estimate changes, which is computationally expensive. This gap drags the application of SDF in highly-dynamic scenarios.

To apply the SDF, two possible solutions are proposed. In the first one, the global SDF is required to be updated every time the pose estimate changes. However, the update may be computationally expensive when the environment is complex and highly-dynamic. In the second one, the SDFs is stored in a neural network by utilizing learning approaches. For example, \cite{2024_ICRA_dynamic_neural_fields_for_robot_self-modeling}, \cite{2022_SR_Fully_body_visual_self_modeling} presented a full-body visual self-modeling approach that captures the entire robot morphology and kinematics using a single implicit neural representation, i.e., a global SDF to answer queries as to whether a spatial position (x,y,z) will be occupied if the joints move to some specified angles. However, this solution requires a tedious and costly offline stage to collect extensive labeled data for a specific scenarios. Thus it is difficult to be applied in highly-dynamic scenarios where the scene changes frequently.

In this paper we present a chained version of SDFs for real-time reflexes in highly-dynamic scenarios. Different to existing solutions that aims to update the global SDF every time or to store the SDFs in a neural network, we propose a model-based way to store a series of local SDFs for describing the geometric information of the static environment, the geometric information of each link, and the reachability of the manipulator, respectively, in the offline stage. These local SDFs can be pre-computed and only need to be computed once. In addition, for different objects, their SDFs can choose different resolutions to discretize the space. In the online stage, these local SDFs are chained together according to the configuration of the manipulator, to provide global geometric information about the manipulator and the environment. Then these SDFs serve as a lookup table for dynamic objects to quickly generate real-time reflex behaviors to avoid the static and dynamic obstacles emerging in the environment. Compared with existing solutions, our chained version of SDFs avoids the update process and can be easily adapted to different scenes. Our main contributions are summarized as:
\begin{itemize}
    \item  We design an unified implicit representation framework for describing the geometric information of the manipulator and its working environment in a model-based way. To achieve this, we create 3 groups of local SDFs in the offline stage. The first ones are to describe the geometric information of the static environment. The second ones are to describe the geometric information of each link. The third ones are to describe the reachability of each link's end of the manipulator. All these local SDFs can be pre-computed and stored. When the environment is changed, we only need to update the first group of SDFs, making the whole system be highly flexible and extendable.
    \item We propose an efficient algorithm to chain the local SDFs together in the online stage for providing global geometric information about the manipulator and its environment. According to the configuration of the manipulator, these local SDFs serves as lookup tables for the point clouds of the dynamic objects to query, for generating real-time escape velocities according to the distance and gradient information. 
    \item We propose a modified geometric Jacobian matrix and then use the Jacobian-pseudo-inverse method to generate real-time reflex behaviors. During this stage, the escape velocity drive the manipulator to take actions to avoid the static and dynamic obstacles emerging in the environment, which is similar to the unconditioned reflex of humans.
    \item Experimental comparisons with existing solutions are designed and conducted in both static and dynamic scenarios. In the static scenario, our method identifies the path solutions with lower time consumption and shorter trajectory length. In the dynamic scenario, our method can reliably pursue the dynamic target point, avoid dynamic obstacles, and react to these obstacles within 1 ms, which surpasses the unconditioned reflex reaction time of humans.
\end{itemize}

This paper is organized as follows. After discussing related work in Sect. 2, we will introduce the mathematical preliminaries and problem definition in Sect. 3. Then our implicit representation framework by using local SDFs and the unconditioned reflex mechanism are described in Sect. 4 and Sect. 5, respectively. Simulation and Experimental evaluations are given in Sect. 6. Finally, Sect. 7 gives our discussion and Sect. 8 concludes the paper.

\section{2. Related Work}
\label{Section2_RelatedWork}

In this chapter, we would like to briefly review the related work on path/trajectory planning algorithms.

The conventional popular path planning algorithms are mainly divided into three types: the sampling-based ones, the graph-based searching ones, and the optimization-based ones.

In the realm of sampling-based path planning algorithms, the most influential one to date remains the Probabilistic RoadMaps (PRM) (\cite{1996_TRO_PRM}). The PRM algorithm, as a multiple-query method, first constructs a comprehensive graph representing a rich set of collision-free trajectories by sampling in the configuration space and performing collision detection on the sampled points. It then obtains feasible paths through graph search. The advantages of PRM lies in its efficiency, which depends primarily on the difficulty of finding paths rather than the dimensionality of the configuration space. Therefore, it performs well even in high-dimensional spaces. Additionally, the probability of planning failure decreases exponentially to zero as the number of sampled points increases (\cite{1998_TRO_PRM}). Hence, the PRM algorithm is probabilistically complete.

However, when a robot's working environment is unknown in advance or when the robot frequently switches between different environments, the multiple-query methods become less effective. Therefore, the Incremental Sampling-based Rapidly-exploring Random Tree (RRT) algorithm (\cite{2001_IJRR_RRT}) and its asymptotically optimal variant, RRT* (\cite{2011_IJRR_Sampling_based_algorithms}), were primarily proposed for unstructured environments. The RRT* algorithm builds a search tree of valid paths rooted at the start by incrementally sampling and connecting states when these local connections are valid. This approach avoids the need to predefine the number of sampled points while also constraining the sampling region. Additionally, the RRT* algorithm had been proven to be probabilistically complete. However, because the sample points are generated randomly, the RRT* algorithm cannot guarantee that the final path is globally optimal, particularly in complex scenarios.

Different to sampling-based methods, the graph-based searching methods can consistently provide optimal solutions on a global scale. One of the most common graph searching algorithms is the Dijkstra's algorithm (\cite{1959_NM_Dijkstra}). Its main idea is to approach a target point from an initial point by employing a greedy strategy. During each traversal, it selects the nearest neighboring node to the starting point that has not been visited yet, continuing until it reaches the destination. Dijkstra's algorithm primarily addresses the problem of finding the shortest path in weighted graphs. However, as the dimensionality of the search space increases, its searching efficiency decreases exponentially.

Currently, there is a method that combines graph searching techniques with sampling methods. It leverages the heuristic searching from the A* algorithm (\cite{1968_TSSC_Astar}) and the informed sampling-based incremental sampling from the RRT* algorithm. This method is known as the Batch Informed Trees (BIT*) algorithm (\cite{2020_IJRR_BIT}). The BIT* algorithm employs a multi-batch, multi-sampling method, allowing it to quickly return feasible solutions before finding the optimal solution during the searching process. Additionally, within each sampling batch, it introduces heuristic values to perform ordered searches near the optimal solution, thereby enhancing searching efficiency.

But both the graph-based searching and the sampling-based methods require a step-by-step expansion of nodes in the configuration space. No matter expanding a single node or a batch of nodes in a single iteration, there is no guarantee that any of the previously expanded nodes are valid until the target point is reached. The manipulator can only follow the desired trajectory after the entire planning algorithm is completed. But dynamic obstacles can disrupt the generated path at any time. Consequently, these motion planning approaches are inherently not suitable for dynamic scenarios.

Besides graph-based searching and sampling-based methods, the optimization-based motion planning is also a prevalent approach. The main challenge of optimization-based methods lies in transforming constraints from task space to configuration space. Currently, there are two approaches to address this difficulty. The first method attempts to transform the obstacles from task space to configuration space to obtain an unreachable region, by using inverse kinematics. The unreachable region serves as constraints for the optimization problem. However, these constraints are generally hard to be described by analytical expressions, making it very difficult to solve the optimization problem (\cite{1988_canny_complexity}). In addition, this method can only handle obstacle avoidance for the end-effector part. The collision with the other parts of the manipulator, e.g., the links, are difficult to be considered by this method. Despite the aforementioned shortcomings, some pleasant results have been achieved. It has been demonstrated (\cite{2012_latombe_robot_motion_planning}) that if all constraints in the task space can be described as semialgebraic sets, which are the intersections and unions of polynomial inequalities, then the constraint set in the configuration space is also a semialgebraic set. This indicates that although the method is currently challenging, it is solvable.

The second method attempts to directly describe the reachable regions in configuration space as a union of several simple sets. The IRIS algorithm (\cite{2015_STAR_Computing_Large_Convex_Regions}) achieved this by decomposing obstacles in known, convex configuration spaces into a set of reachable space regions, which are convex. Building upon this, some scholars (\cite{2023_IJRR_Certified_polyhedral_decompositions}) proposed a method to describe the reachable regions using convex polyhedra in a bijective, rational parametrization of the configuration space, which is known as the tangent configuration space. This method efficiently generates reachable space in configuration space by decomposing it into multiple convex polyhedra. However, such methods are still primarily applicable for solving path planning problems in static, known environments. The trajectory planning with dynamic obstacles remains challenging.

Building upon convex optimization methods, a Convex Set Graph (GCS) trajectory optimization algorithm framework (\cite{2023_SR_Motion_planning_around_obstacles_with_convex_optimization}), that integrates graph searching with convex optimization, was proposed. This framework facilitates efficient and reliable trajectory planning around obstacles using convex optimization techniques. By leveraging GCS methods for finding shortest paths, the algorithm incorporates a convex relaxation based on linear programming (LP) or second-order cone programming (SOCP), coupled with a randomized rounding algorithm. This approach ensures the determination of a globally optimal collision-free path. However, it is noteworthy that this method is applicable only to static obstacle scenarios.

In recent years, with the development of commercial solvers, the optimization-based path planning methods have gradually achieved very low time consumption and high success rates. However, most of these successful outcomes require the algorithms to pre-compute the current working environment during the offline stage, and the manipulator must wait for the solver to finish solving the objective function before it can react. Therefore, the optimization-based path planning methods are not suitable for highly-dynamic environments.

Except for the sampling-based, the graph-based searching, and the optimization-based methods, another group of related work is on the local re-planning strategy, which was proposed to address dynamic obstacles. A common approach is to utilize the RRT algorithm and its variants for local replanning (\cite{2006_ICRA_Replanning_with_rrts}, \cite{2021_RSJ_Robotic_lime_picking_by_considering}). For a path temporarily blocked by a dynamic obstacle, \cite{2023_RAL_Path_Re_Planning_Design} proposed to delete the nodes that detected collisions and replan using the APF and RRT with the current point as the root node and any point in the original path after the collision as the connecting target. Similarly, \cite{2013_ASCC_VF_RRT} proposed an RRT-based algorithm for re-sampling using a potential field, called Vector Field RRT (VF-RRT), which dynamically balances the weights of the random sampling and the potential energy sampling based on the space occupied by obstacles. This method preserved the benefits of paths based on the potential energy sampling while avoiding local minima. However, efforts to adapt sampling-based methods for dynamic environments encounter significant challenges. These adaptations often either perform worse than the original sampling-based approaches in terms of time consumption and path quality, or fail to achieve effective dynamic obstacle avoidance. These phenomena validated that these methods are inherently not suitable for dynamic scenarios, because, for instance, they might only achieve end-effector avoidance without addressing real-time obstacle dynamics effectively.

Recently, besides using re-planning methods, a method based on the modulated dynamic system (DS) (\cite{2012_AR_DS}) has been proposed for reactive motion planning (\cite{2024_IJRR_ReactiveMPC}). This method defines a vector field to guide the motion of the manipulator while employing a dynamic modulation matrix to update the vector field in real time. Furthermore, a reactive collision-free motion generation method which combines the dynamic modulation system with sampling-based Model Predictive Control (MPC) has been proposed. By optimizing the real-time dynamic modulation matrix through sampling-based MPC, this approach ensures that the vector field does not fall into local minima near non-convex obstacles. However, the calculation of the dynamic modulation matrix for complex and highly-dynamic environments is still a bottle-neck for real-time update, leading to a only 20Hz frequency loop, which is fall behind the unconditioned reflex reaction time of humans.

The last group of related work is about the SDF-based motion planning algorithms. As stated before, the aforementioned approaches are all based on explicit representation methods, in which a time-consuming searching for the obstacle avoidance velocity is inevitable. The time-consuming searching makes it impossible to generate real-time reflexes for highly-dynamic events. To overcome the time-consuming issue of explicit methods, the SDF is introduced since it can replace the time-consuming searching by a highly-efficient querying on the pre-computed SDF table at runtime. However, the articulated objects, e.g., manipulators, bring challenges to the pre-computation of the SDF.

To bridge this gap, two possible solutions are proposed. In the first one, the global SDF is defined over the space instead of different objects. In this way, the space is discretized into a group of voxels and the SDF is updated over each voxel by considering all objects. For example, \cite{2016_ICRA_Considering_avoidance_and_consistency_in_motion_planning} introduced two occupancy grids over the space, to record occupancy over time for both a robot and human collaborator.
Furthermore, \cite{2021_PICAPS_Predicted_composite_signed-distance_fields} demonstrated that the global SDF value over a specific voxel can be superposed by all the object SDFs using a $min$ operation, resulting in a composite SDF.
This solution avoids the pre-computation for articulated objects. Instead, it requires to update the SDF values over all voxels in the space. For a space with sparse objects, this update is faster than the collision check by utilizing explicit representation methods. Thus this solution has achieved success in human-robot collaboration and real-time motion planning in dynamic environments.
However, this solution brings dimensional explosion problem. It is difficult to tune a proper resolution to discretize the space. To guarantee the obstacle avoidance for all objects, the resolution has to be chosen as the highest one. With the improvement of precision requirements, the computation time of the global SDF update is greatly increased.

The other solution is to store the SDFs in a neural network by utilizing learning approaches. For example, \cite{2024_ICRA_dynamic_neural_fields_for_robot_self-modeling}, \cite{2022_SR_Fully_body_visual_self_modeling} presented a full-body visual self-modeling approach that captures the entire robot morphology and kinematics using a single implicit neural representation, i.e., a global SDF to answer queries as to whether a spatial position $\left(x, y, z\right)$ will be occupied if the joints move to some specified angles. Similarly, \cite{2024_arxiv_CDF} proposed to plan in the robot configuration space, in which a neural Configuration-space Distance Filed (CDF) representation is introduced and computed by using Multilayer Perceptrons (MLPs).
Another two approaches were proposed by \cite{2022_RAL_Neural_joint_space_implicit} and \cite{2022_RSJ_Regularized_deep_signed_distance_fields}, respectively. \cite{2022_RAL_Neural_joint_space_implicit} utilized model files such as URDF to collect distance data from points in space relative to the surface of the manipulation under different configurations, generating and storing a state-indexed full-body SDF for the manipulation within a neural network. \cite{2022_RSJ_Regularized_deep_signed_distance_fields} utilized point cloud data from the manipulation in various configurations. By expanding outward along the normal direction of each surface point on the object, they obtained a series of distance contours. These contour points were then used as training data to generate a neural network that models the full-body SDF of the manipulation. However, only the manipulator is modeled and the information about the dynamic environment is missing in these two approaches.
By using learning approaches, all the signed distanced functions, which are indexed by states, are generated and stored in a neural network. When a query is requested, the current states, e.g., the joint angles of the manipulator, together with the spatial locations, are provided as inputs to the neural network to get the occupancy information.
However, the learning approaches exhibit poor interpretability and generalizability. Generally, this solution requires a tedious and costly offline stage to collect extensive labeled data for a specific scenarios. When the robot faces a new scene, the data collection and the training process require to redo to obtain a new neural network. Thus it is difficult to be applied in highly-dynamic scenarios where the scene changes frequently.

\section{3. Mathematical Preliminaries}
\label{Section3_ProblemStatement}

The Signed Distance Function (SDF) is a level set function that describes the distance between a given point and the boundary of a set $\Omega$ in a metric space. For better illustration of implicit methods, an object $\Omega$ with irregular shape is taken as an example.

\begin{figure}[ht]
    \centering
    \includegraphics[scale = 0.25]{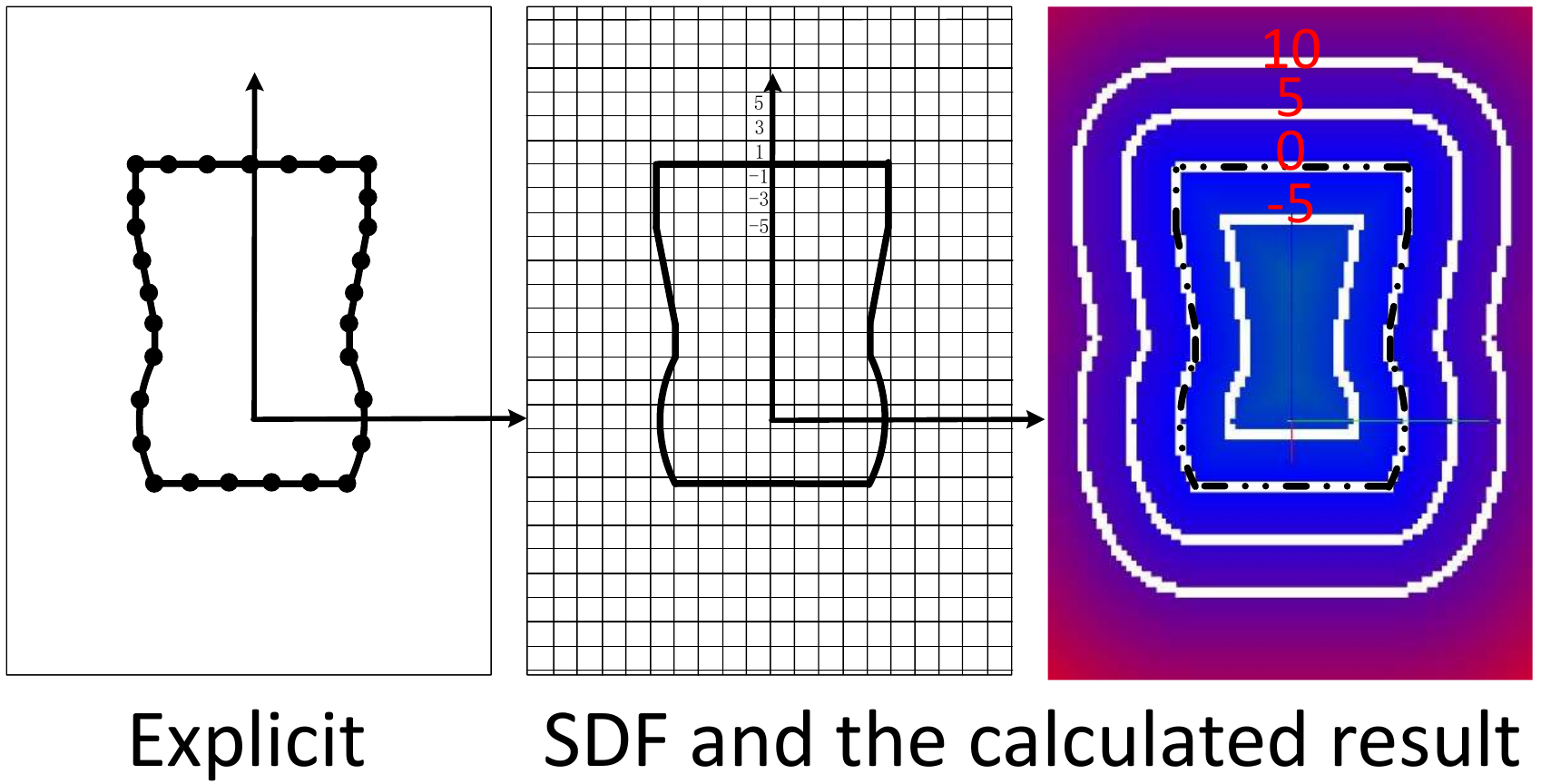}
    \caption{The difference between the explicit method and the SDF method is presented. The explicit method represents an object with multiple points and the expressions of functions between these points. In contrast, the SDF returns the signed distance from any point in space to the surface of the $\Omega$. The dashed lines on the right are contour lines that indicate the distance from the surface of the $\Omega$.}
    \label{SDF_intro}
\end{figure}

As shown in the left of Fig. \ref{SDF_intro}, the geometry of the object $\Omega$ is stored as a set of points/lines/triangles for explicit representation methods. However, to achieve obstacle avoidance, the computation of the nearest distance between any two objects is necessary, but has to depend on a time-consuming searching for complex and irregular curves/surfaces/objects. Therefore, based on explicit representation methods, it is super difficult to enable the unconditioned reflexes for robots, especially in highly-dynamic scenarios.

Different to explicit methods, the SDF, which is an implicit method, is defined as a level set function over the whole space. As shown in the right of Fig. \ref{SDF_intro}, the SDF value is assigned as 0 for points on the surface, positive distance for points on the exteriors, and negative distance for points on the interiors of the object $\Omega$:

\begin{align}
    \label{Eq_SDF_definition}
    \begin{split}
        SDF({\vec p})=\begin{cases}
            d({\vec p}, \delta\Omega)\;,& \mbox{if } {\vec p} \in \Omega^+\\
            0\;,& \mbox{if } {\vec p} \in \delta\Omega\\
            -d({\vec p}, \delta\Omega)\;,& \mbox{if } {\vec p} \in \Omega^-
        \end{cases}
    \end{split}
\end{align}
where ${\vec p}$ is the given point in the space, $\Omega^-$ refers to the interior of the object, $\Omega^+$ refers to the exterior of the object, $\delta\Omega$ refers to the surface of the object. $d$ takes the Euclidean distance function in piecewise. In addition, except for the distance information, the gradient information and/or the parent points (the corresponding points on the surface $\delta\Omega$ that has the nearest distance to position $\vec p$) can also be pre-computed and stored according to needs.

Based on SDF, the geometric information of the object $\Omega$ is embedded in the zero values. The escape velocity to avoid the obstacle $\Omega$ can be easily obtained by a simple and extremely fast lookup from the stored SDF values. However, the limitation of the SDF is also obvious. First, the calculation of the SDF is time-consuming. Therefore, the SDF can only be used to describe rigid-body objects that can be pre-computed. For articulated objects, e.g., manipulators, the pre-computation is challenging, because the articulated models results in a continuous space of signed distanced functions, as a function of the state. How to overcome this limitation and apply the SDF to the motion planning problem of manipulators in highly-dynamic scenarios is our main goal in this paper.

\begin{figure}[ht]
    \centering
    \includegraphics[scale = 0.55]{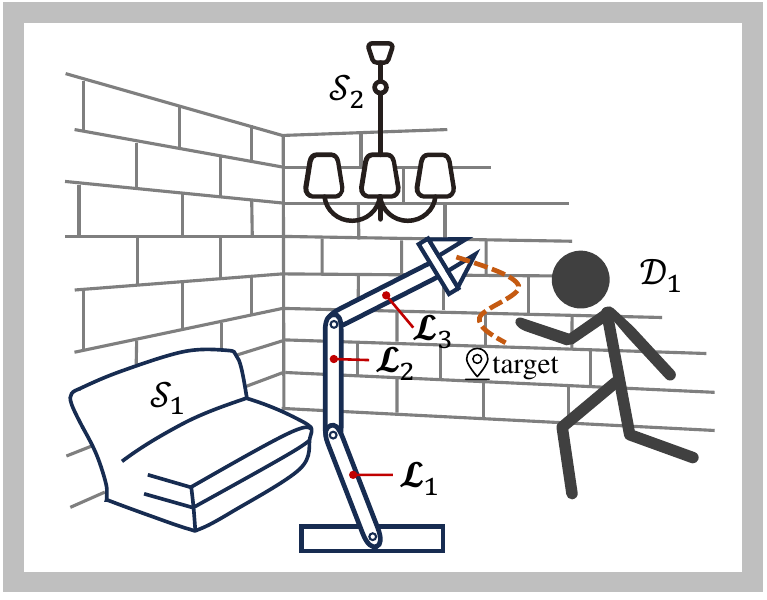}
    \caption{Illustration of a generalist robot working in a household environment, featuring both static and dynamic obstacles. The manipulator must respond in real time to avoid these obstacles and reach the target pose.}
    \label{problem_statement}
\end{figure}

Consider the motion planning problem for a manipulator working in a highly-dynamic environment where both static and dynamic obstacles exist. For better illustration, the following notations are defined:
\begin{itemize}
    \item $\{W\}, \{B\}$: The world coordinate system and the base coordinate system of the manipulator. Without loss of generality, we assume that the world coordinate system $\{W\}$ is aligned with the base coordinate system $\{B\}$ of the manipulator. In this paper, we would use the left superscript to indicate in which coordinate system a variable is expressed. Without special instructions, the variable is expressed in the world frame $\{W\}$.
    \item $\mathcal{M} := \left\{\mathcal{L}_1, \mathcal{L}_2, \cdots, \mathcal{L}_N\right\}$: The manipulator which is composed by $N$ links. $\mathcal{L}_i$ represents the $i$-th link and $\{L_i\}$ is the coordinated system fixed on $\mathcal{L}_i$. Here $i \in \left[1, N\right]$, where $N$ is the Degrees-of-Freedoms (DoFs) of the manipulator. The shapes of each link can be obtained from the CAD model or a URDF/SRDF file of the manipulator. All the links are assumed to be rigid-bodies.
    \item $\mathcal{S} := \left\{\mathcal{S}_0, \mathcal{S}_1, \mathcal{S}_2, \cdots, \mathcal{S}_{M-1}\right\}$: The set of static objects in the working environment, where $\mathcal{S}_i$ ($i \in [1, M]$) represents the $i$-th static object and $M$ is the number of the static objects. In this paper, an object is claimed to be static if its position and orientation keeps constant relative to the world frame $\{W\}$. All the static objects are assumed to be rigid-bodies. Interchangeably, we would use $\left\{\mathcal{S}_i\right\}$ refer to the frame fixed on the object $\mathcal{S}_i$.
    \item $\mathcal{D}\left(t\right) := \left\{{}^W{\vec p}_1\left(t\right), {}^W{\vec p}_2\left(t\right), \cdots, {}^W{\vec p}_K\left(t\right)\right\}$: The point cloud of the dynamic objects at time $t$, which are represented by a series of points ${}^W{\vec p}_i \in \mathbb{R}^3$ ($i \in \left[1, K\right]$). $K$ is the size of the point cloud $\mathcal{D}\left(t\right)$. For dynamic objects with known shape, $\mathcal{D}\left(t\right)$ can be extracted from its SDF, i.e., the set of voxels with SDF values being zero. For unpredictable dynamic objects, it is impossible to pre-known its SDF. But the point cloud can be perceived by sensors like RGB-D camera or Lidar. Please note that the point cloud $\mathcal{D}\left(t\right)$ is not limited to being generated by a single object, but can also be a combination of point clouds generated by multiple objects.
    \item ${\vec q}\left(t\right) \in \mathcal{Q} \subset \mathbb{R}^N$: The joint states of the manipulator at time $t$, where $\mathcal{Q}$ is the $N$-dimensional configuration space of the manipulator.
    \item ${}^B{\vec p}\left(t\right) \in \mathbb{R}^3$: The Cartesian position of the manipulator at time $t$, expressed in the base frame \{B\}.
    \item ${}^B{\mat R}\left(t\right) \in SO\left(3\right)$: The Cartesian orientation matrix of the manipulator at time $t$, expressed in the base frame \{B\}.
    \item ${}^W{\vec p}_{tar} \in \mathbb{R}^3$: The target position for the manipulator, expressed in the world frame \{W\}.
    \item ${}^W{\mat R}_{tar} \in SO\left(3\right)$: The target orientation for the manipulator, expressed in the world frame \{W\}.
    \item ${}^W{\vec p}_{\mathcal{S}_i} \in \mathbb{R}^3, {}^W{\mat R}_{\mathcal{S}_i} \in SO\left(3\right)$: The Cartesian position and orientation of the object $\mathcal{S}_i$, expressed in the world frame \{W\}.
\end{itemize}

To make it more clear, the following assumption is made:
\begin{assumption} \label{Assum-static-map-be-known}
The shapes and poses ($\left[{}^W{\vec p}_{\mathcal{S}_i}, {}^W{\mat R}_{\mathcal{S}_i}\right]$) of all the static objects can be known from a CAD model or the whole static environment can be measured by utilizing 3D scene reconstruction algorithm.
\end{assumption}





Based on the aforementioned notations and assumption, our problems can be illustrated as:

\begin{problem} \label{Prob-RepresentMethod}
Given the manipulator system $\mathcal{M}$ with its joint states ${\vec q}\left(t\right)$, the static object set $\mathcal{S}$, the point cloud $\mathcal{D}\left(t\right)$ of the dynamic objects, design an implicit representation framework such that all the SDFs can be pre-computed and stored, while the distance and gradient information to obstacles can be extracted in sub-milliseconds.
\end{problem}

\begin{problem} \label{Prob-UnconditionedReflex}
Given the manipulator system $\mathcal{M}$ with its joint states ${\vec q}\left(t\right)$ and the implicit representation framework given by solving {\bf Problem \ref{Prob-RepresentMethod}}, design an unconditioned reflex mechanism such that the manipulator can generate the escape velocity ${\vec \upsilon}\left(t\right)$ in real-time and take actions ${\vec q}\left(t\right)$ to avoid obstacles immediately (in milliseconds), while the target pose $\left[{\vec p}_{tar}, {\mat R}_{tar}\right]$ can finally be reached, i.e.,
\begin{align}
    \label{Eq_target}
    \begin{split}
        \lim_{t \to \infty} {\left\| {\vec p}_{tar} - {\vec p}\left(t\right) \right\| \leq \epsilon_p}, \\
        \lim_{t \to \infty} {\left\| \log\left({\mat R}^\top\left(t\right) {\mat R}_{tar}\right)\right\| \leq \epsilon_r},
    \end{split}
\end{align}
where $\epsilon_p$ and $\epsilon_r$ are small constant values to indicate the tracking accuracy, which are determined by users.
\end{problem}

\section{4. The Implicit Framework: Geometric Representation of the Manipulator and its Working Environment Using local SDFs}
\label{Section5_Method}

Different from conventional methods that define a global SDF over the whole space or store the global SDF in a neural network, our main idea is to pre-compute and store some local SDFs in the offline stage. In the online stage, these local SDFs can be chained together to provide global geometric information about the manipulator and its environment. In this section, we would like to provide a detailed description of the proposed local SDFs, including its construction and how these local SDFs can be chained to represent the entire system.

 In order to provide a comprehensive understanding, we classify the objects in the environment into 4 levels:
\begin{itemize}
    \item {\bf Level 1: Permanent Objects}. In the working environment, many objects, e.g., the walls of a building, are permanently static. All these permanent objects can be unified as a single object, denoted as $\mathcal{S}_0$ in this paper. For $\mathcal{S}_0$, we can define a unified SDF fixed to the world frame $\{W\}$ to describe the occupancy information. This SDF can be constructed from the architectural blueprints of the building, or from the 3D reconstruction result via point clouds scanned by sensors like Lidar. Please not that this construction only require to do once for a specific environment.
    \item {\bf Level 2: Near-Permanent Objects}. Many objects, e.g., furnitures like fixed table, refrigerator, and wardrobe, can be considered as immobile in a short-term period. However, the layout can be adjusted when the house is refurnished. These objects, denoted as $\mathcal{S}_0, \cdots, \mathcal{S}_{M-1}$, are named as near-permanent objects in this paper. To guarantee better flexibility, we can create a local SDF fixed on each near-permanent object to describe its shape. These local SDFs, together with their poses $\left[{}^W{\vec p}_{\mathcal{S}_i}, {}^W{\mat R}_{\mathcal{S}_i}\right]$, can completely describe the geometric information of these objects. When the layout is changed, we only need to update their poses $\left[{}^W{\vec p}_{\mathcal{S}_i}, {}^W{\mat R}_{\mathcal{S}_i}\right]$. When a new furniture $\mathcal{S}_M$ is added into the environment, we can simply add the new SDF together with its pose $\left[{}^W{\vec p}_{\mathcal{S}_M}, {}^W{\mat R}_{\mathcal{S}_M}\right]$ to the static object set $\mathcal{S}$. Therefore, our design make the whole system be highly flexible and extendable. Another benefit to separate the near-permanent objects from the permanent objects is that, we can use different spatial resolutions to define the local SDFs for different objects, according to needs.
    \item {\bf Level 3: The Manipulator}. The manipulator $\mathcal{M}$, which is composed of $N$ serial links, is a typical articulated object. Although its shape changes over time, the shape of each link can be regarded as unchanged once the manipulator is manufactured. Therefore, with knowing the shapes of all links, the global shape of the manipulator can be identified by its configuration (i.e., joint states). Hence we can define a group of local SDFs fixed on each link frame $\mathcal{L}_i$, together with its joint states ${\vec q}\left(t\right)$, to store the geometric information of the manipulator $\mathcal{M}$. The manipulator $\mathcal{M}$ can either be a fixed-base one or a floating-base one. For a fixed-base manipulator, the local SDFs, together with its joint states, are enough to describe its status. For a mobile manipulator, the pose of the base can be added to describe its status. In this paper, we would limit the manipulator to be a fixed-base one. But our proposed implicit framework can also handle the floating-base ones.
    \item {\bf Level 4: Dynamic Objects}. Different to constructed environment, the dynamic objects appear and disappear frequently in unstructured environment, bring significant challenges to the motion planning problem. In this paper, we would regard all objects that is movable during the online stage as dynamic objects, even though its current velocity is zero. For example, objects waiting to be operated on a table, is modeled as dynamic objects, even if they keeps static during the motion of the manipulator. For these objects, their shapes can be pre-known and described by a local SDF fixed on themselves. With known its pose, the point cloud $\mathcal{D}\left(t\right)$ of this object can be extracted from its SDF, i.e., the set of voxels with zero value. For unpredictable objects, it is difficult to pre-know their shapes. However, their occupancy information $\mathcal{D}\left(t\right)$ can be sensed by sensors like RGB-D camera or Lidar.
\end{itemize}

Based on the above features, we define 3 groups of local SDFs, which are all able to be pre-computed and stored, to store the geometric information of the manipulator and its working environments. Among which, one group is to describe the geometry of the permanent and near-permanent objects and the other twos are for the manipulators. While the points in $\mathcal{D}\left(t\right)$ serve as query point for the local SDFs. These local SDFs are detailed as follows.

{\bf (1) Local SDFs to describe the shapes of the permanent and near-permanent objects $\mathcal{S}_i$:}

\begin{figure}[ht]
    \centering
    \subfigure[]{\includegraphics[width = 0.48\hsize]{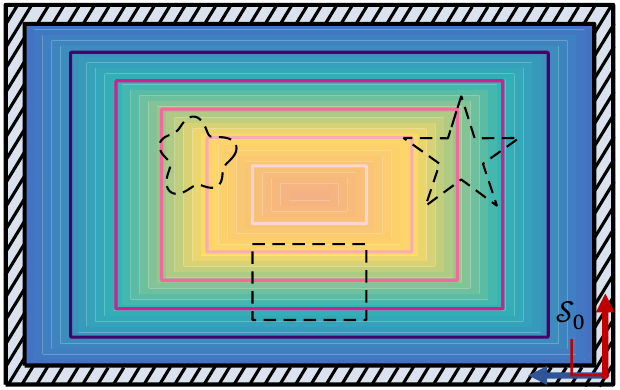}\label{Fig_Si_intro_0}}
    \hspace{0.01\hsize}
    \subfigure[]{\includegraphics[width = 0.48\hsize]{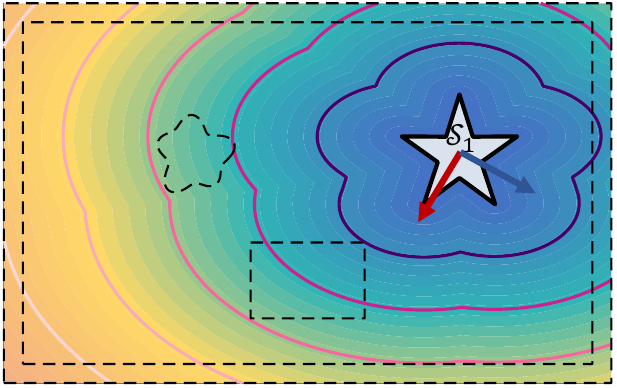}\label{Fig_Si_intro_1}}
    \vspace{0.001\hsize}

    \subfigure[]{\includegraphics[width = 0.48\hsize]{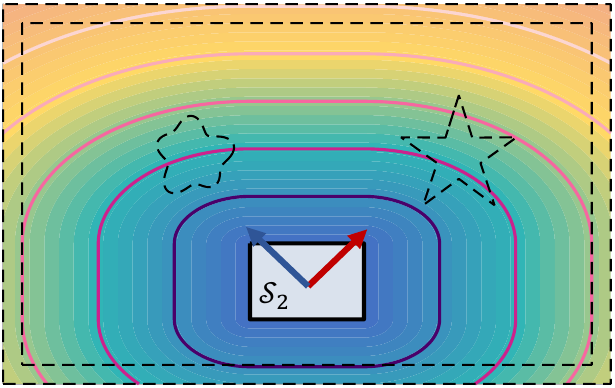}\label{Fig_Si_intro_2}}
    \hspace{0.001\hsize}
    \subfigure[]{\includegraphics[width = 0.48\hsize]{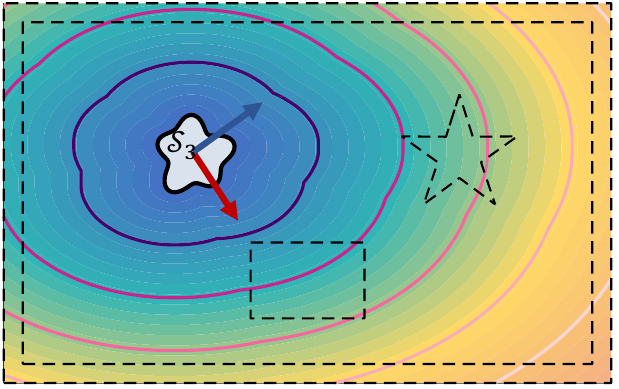}\label{Fig_Si_intro_3}}
    \vspace{0.01\hsize}
    \subfigure[]{\includegraphics[width = 0.99\hsize]{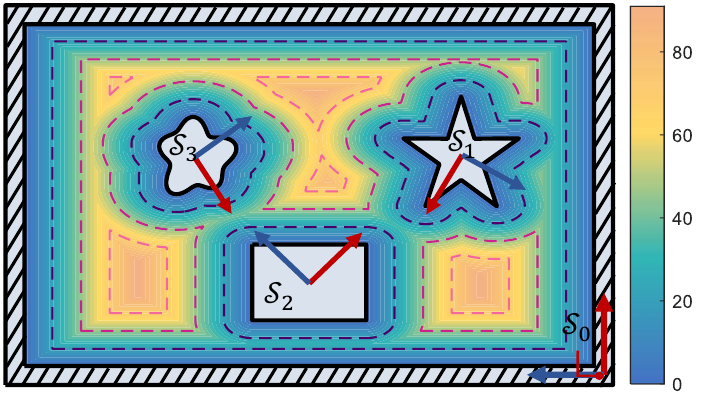}\label{Fig_Si_intro_4}}

    \caption{(a) depict the $SDF_{\mathcal{S}_0}\left({\vec p}\right)$  created for the permanent object, e.g., the wall. (b), (c) and (d) depict the schematic diagram of the $SDF_{\mathcal{S}_i}\left({\vec p}\right)$ created for each static obstacle. (d) shows the $SDF_{\mathcal{S}}\left({\vec p}\right)$, which is the fusion of all the local $SDF_{\mathcal{S}_i}\left({\vec p}\right)$. The colors indicate the signed distance (Unit: mm). The contour lines are also marked in each subfigure.  }
    \label{Fig_Si_intro}
\end{figure}

\begin{itemize}
    \item $SDF_{\mathcal{S}_i}\left({\vec p}\right)$: The local SDF that describes the geometric information of the $i$-th static obstacle $\mathcal{S}_i$. It returns the signed distance to the surface of $\mathcal{S}_i$ for any query point ${}^{\mathcal{S}_i}{\vec p} \in \mathbb{R}^3$ that defined in the frame $\{\mathcal{S}_i\}$.

    \item $SDF_{\mathcal{S}}\left({\vec p}\right)$: The SDF that describes the geometric information of all static obstacles throughout the workspace. It returns the shortest signed distance that a given point to the surfaces of all the static objects. As illustrated in Fig. \ref{Fig_Si_intro}, $SDF_{\mathcal{S}}\left({\vec p}\right)$, which is defined in the world frame $\{W\}$, is obtained by compositing all $SDF_{\mathcal{S}_i}\left({\vec p}\right)$. For a specific scene, this composition only need to do once. When a new static object ${\mathcal{S}_M}$ is added into the environment or the pose of a static object ${\mathcal{S}_i}$ is changed, we only need to add the new SDF function $SDF_{\mathcal{S}_M}\left({\vec p}\right)$ (together with its pose $\left[{}^W{\vec p}_{\mathcal{S}_M}, {}^W{\mat R}_{\mathcal{S}_M}\right]$) or update the pose $\left[{}^W{\vec p}_{\mathcal{S}_i}, {}^W{\mat R}_{\mathcal{S}_i}\right]$, and then re-calculate the $SDF_{\mathcal{S}}\left({\vec p}\right)$ once.
\end{itemize}

{\bf (2) Local SDFs to describe the links' shapes of the manipulator $\mathcal{M}$:}

\begin{figure}[ht]
    \centering
    \subfigure[]{\includegraphics[width = 0.99\hsize]{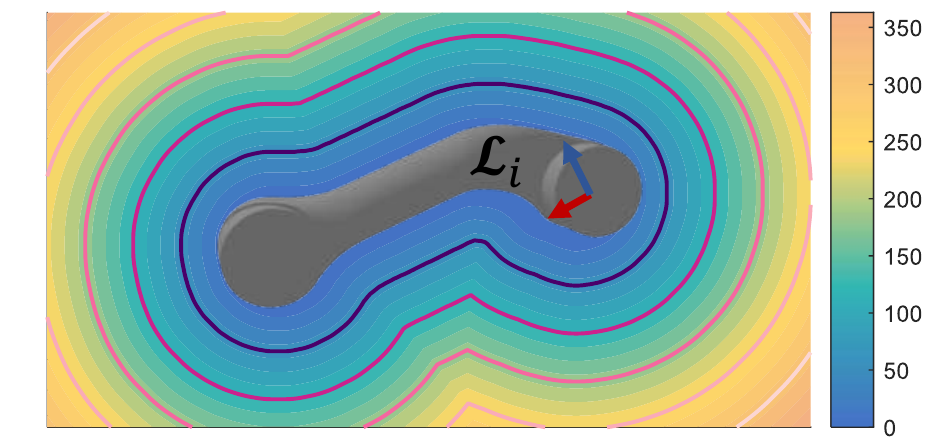}\label{Fig_Li_intro_1}}
    \subfigure[]{\includegraphics[width = 0.99\hsize]{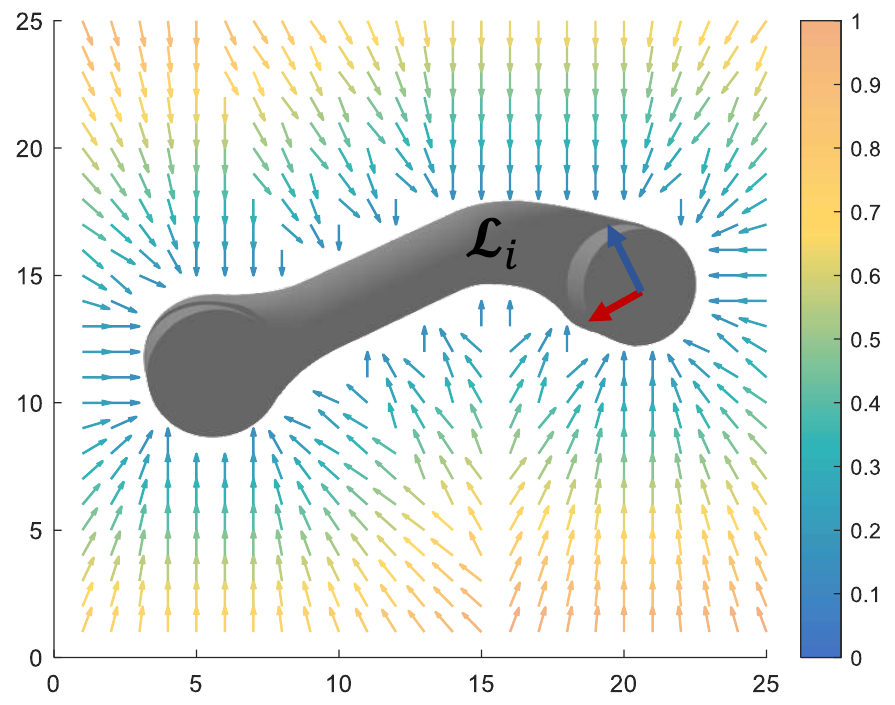}\label{Fig_Li_intro_2}}
    \caption{(a) shows the schematic diagram of the $SDF_{\mathcal{L}_i}\left({\vec p}\right)$ and (b) shows the $Gradient_{\mathcal{L}_i}\left({\vec p}\right)$. These local SDFs returns the signed distance and gradient information to the surface of $\{\mathcal{L}_i\}$. In (a), the colors indicate the signed distance (Unit: mm) while the colored lines mark the contour lines. In (b), the direction and magnitude of the gradient are illustrated by the arrows' direction and color, respectively. The magnitude is normalized for visualization.}
    \label{Fig_Li_intro}
\end{figure}

\begin{itemize}
    \item $SDF_{\mathcal{L}_i}\left({\vec p}\right)$: The local SDF that describes the geometric information of $\mathcal{L}_i$. It returns the signed distance to the surface of $\mathcal{L}_i$ for any query point ${}^{\mathcal{L}_i}{\vec p} \in \mathbb{R}^3$ that defined in the frame $\{\mathcal{L}_i\}$.

    \item $Gradient_{\mathcal{L}_i}\left({\vec p}\right)$: Except for the traditional signed distance information, here we would also pre-compute and store the gradient information $Gradient_{\mathcal{L}_i}\left({\vec p}\right)$ of each link ${\mathcal{L}_i}$, for the convenience of obstacle avoidance. As shown in Fig. \ref{Fig_Li_intro}, the $Gradient_{\mathcal{L}_i}\left({\vec p}\right)$ returns the gradient of the signed distance for any query point ${}^{\mathcal{L}_i}{\vec p} \in \mathbb{R}^3$ that defined in the frame $\{\mathcal{L}_i\}$.
\end{itemize}

{\bf (3) Local SDFs to describe the links' reachability of the manipulator $\mathcal{M}$:}

\begin{figure}[ht]
    \centering
    \includegraphics[width = 0.99\hsize]{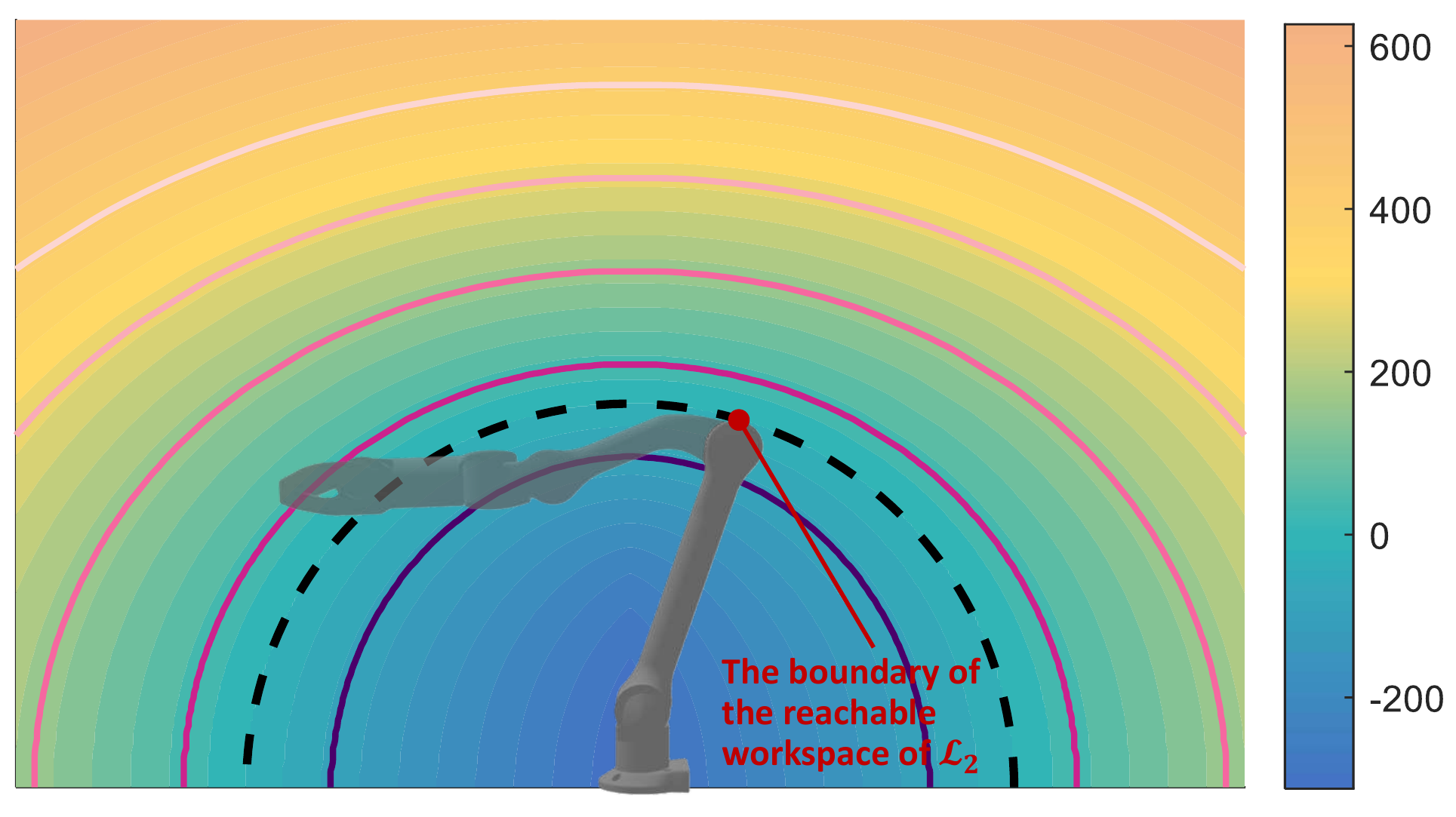}
    \caption{This figure shows $SDF_{{EE}_2}\left({\vec p}\right)$, which describe the reachable workspace of the end of $\mathcal{L}_2$.  }
    \label{sdf_eei_intro}
\end{figure}

\begin{itemize}
    \item $SDF_{EE}\left({\vec p}\right)$: The local SDF to describe the reachable workspace of the manipulator's end-effector. It returns the signed distance to the workspace boundary to indicate the reachability of any query point ${}^B{\vec p} \in \mathbb{R}^3$. We define this function to prune the points which are unreachable for the manipulator, since they are impossible to generate collision with the manipulator.

    \item $SDF_{{EE}_i}\left({\vec p}\right)$: Similar to $SDF_{EE}\left({\vec p}\right)$, we also define the local SDFs to describe the reachable workspace of the end of each link $\mathcal{L}_i$. As shown in Fig. \ref{Fig_Li_intro}, the $SDF_{{EE}_i}\left({\vec p}\right)$ returns the signed distance to the link-end's workspace boundary for any query point ${}^B{\vec p} \in \mathbb{R}^3$. These SDFs can be utilized to further prune the point clouds and accelerate the collision check process. Please note that the $SDF_{EE}\left({\vec p}\right)$ and $SDF_{{EE}_i}\left({\vec p}\right)$ are both defined in the base frame $\{B\}$ of the manipulator.

\end{itemize}

Based on these local SDFs, we construct an implicit representation framework that characterizes both the manipulator and its working environment. Among them, $SDF_{\mathcal{S}_i}\left({\vec p}\right)$ and $SDF_{\mathcal{S}}\left({\vec p}\right)$ describe the geometric information of the static scene. $SDF_{\mathcal{L}_i}\left({\vec p}\right)$ and $Gradient_{\mathcal{L}_i}\left({\vec p}\right)$ compute the manipulator's geometry in detail to facilitate the rapid determination of the shortest distance from any obstacles to the surface of the link, along with the corresponding gradient. $SDF_{EE} $ and $SDF_{\mathcal{L}_i} $ define the reachable workspace of the manipulator, enabling the manipulator to focus on obstacles that present a ``collision risk". Based on all these local SDFs defined in different frames, the geometry of the manipulator and its static working environment is implicitly pre-computed and stored. In the next section, we will explain how to chain these local SDFs in the online stage to generate real-time reflex behaviors.

\subsection{4.1 Construction of $SDF_{\mathcal{S}_i}\left({\vec p}\right)$ and $SDF_{\mathcal{L}_i}\left({\vec p}\right)$ } \label{sec:4.1}

Since the constructions of $SDF_{\mathcal{S}_i}\left({\vec p}\right)$ and $SDF_{\mathcal{L}_i}\left({\vec p}\right)$ are similar, we would like to introduce them together.

Consider an object $\mathcal{S}_i$ or a link $\mathcal{L}_i$, the frame $\left\{\mathcal{S}_i\right\}$ or $\left\{\mathcal{L}_i\right\}$ is fixed on this object. Then for any query point ${}^{\mathcal{S}_i/\mathcal{L}_i}{\vec p}$, which is expressed in $\left\{\mathcal{S}_i\right\}$ or $\left\{\mathcal{L}_i\right\}$, we can pre-compute and store its signed distance to the surface of $\mathcal{S}_i$/$\mathcal{L}_i$. Since it is impractical to store an SDF with infinite resolution and infinite function domain, our first step is to create an enclosure box $E_{\mathcal{S}_i/\mathcal{L}_i}$ that is large enough to cover the given object. Then the enclosure box $E_{\mathcal{S}_i/\mathcal{L}_i}$ is evenly discretized into $N_x \times N_y \times N_z$ voxels, according to a user-defined spatial resolution $\delta_{\mathcal{S}_i/\mathcal{L}_i}$. Please note that the size of $E_{\mathcal{S}_i/\mathcal{L}_i}$ and the resolution $\delta_{\mathcal{S}_i/\mathcal{L}_i}$ to discretize it should be determined according to needs and can be different to each other for different objects.

Then all the voxels in $E_{\mathcal{S}_i/\mathcal{L}_i}$ can be classified into two groups: the ones occupied by the given object and the other ones. The voxels in the occupied group is traversed to check its neighbors. If any of its neighbors is unoccupied, this voxel is regarded to be on the surface of the given object and its SDF value is assigned to be zero. Then all the voxels is traversed again to obtain the signed distance by calculating the Euclidean distance to each boundary voxel. The sign for occupied voxels is negative and the sign for unoccupied ones is positive. The SDF construction algorithm is given in \textbf{Algorithm \ref{algorithm_ConstructionOfSiLi}}.

\begin{algorithm}
\caption{Construction of $SDF_{\mathcal{S}_i/\mathcal{L}_i}\left({\vec p}\right)$}\label{algorithm_ConstructionOfSiLi}
\tcc{To calculate and store the signed distance from any points in the scene to the surface of $\mathcal{S}_i/\mathcal{L}_i$}
\For{$Every\ {\vec v}=(v_i,v_j,v_k)\in E_{\mathcal{S}_i/\mathcal{L}_i}$}{
\If{$E_{\mathcal{S}_i/\mathcal{L}_i}(v_i,v_j,v_k) \;is \;occupied \;$}{
    \If{$E_{\mathcal{S}_i/\mathcal{L}_i}(v_i-1,v_j,v_k) \;is \;unoccupied \; \rVert$\\
        $E_{\mathcal{S}_i/\mathcal{L}_i}(v_i+1,v_j,v_k) \;is \;unoccupied \; \rVert$\\
        $E_{\mathcal{S}_i/\mathcal{L}_i}(v_i,v_j-1,v_k) \;is \;unoccupied \; \rVert$\\
        $E_{\mathcal{S}_i/\mathcal{L}_i}(v_i,v_j+1,v_k) \;is \;unoccupied \; \rVert$\\
        $E_{\mathcal{S}_i/\mathcal{L}_i}(v_i,v_j,v_k-1) \;is \;unoccupied \; \rVert$\\
        $E_{\mathcal{S}_i/\mathcal{L}_i}(v_i,v_j,v_k+1) \;is \;unoccupied \; $\\
    }{
        $SDF_{\mathcal{S}_i/\mathcal{L}_i}(v_i,v_j,v_k)=0$\;
        $\mathcal{B}\leftarrow {\vec v}$ \Comment{$\mathcal{B}$ is the boundary of $\mathcal{S}_i/\mathcal{L}_i$ }
    }
}
}
\For{$Every\ {\vec v}=(v_i,v_j,v_k)\in SDF_{\mathcal{S}_i/\mathcal{L}_i}$}{

    $d_{min} = \overline{d}$\;

    \For{$Every\ {\vec v}_b=(v_i,v_j,v_k)\in B$}{
        \If{$\delta_{\mathcal{S}_i/\mathcal{L}_i}  \lVert {\vec v}-{\vec v}_b \rVert < d_{min}$}{
            $d_{min}=\delta_{\mathcal{S}_i/\mathcal{L}_i}  \lVert {\vec v}-{\vec v}_b \rVert$\;
        }
    }
    \eIf{$E_{\mathcal{S}_i/\mathcal{L}_i}(v_i,v_j,v_k) \;is \;occupied \; $}{
        $SDF_{\mathcal{S}_i/\mathcal{L}_i}(v_i,v_j,v_k)=-d_{min}$\;
    }
    {
        $SDF_{\mathcal{S}_i/\mathcal{L}_i}(v_i,v_j,v_k)=d_{min}$\;
    }
}

\end{algorithm}

\subsection{ 4.2 Construction of $SDF_{\mathcal{S}}\left({\vec p}\right)$: The Composition of all $SDF_{\mathcal{S}_i}\left({\vec p}\right)$ } \label{sec:4.2}

Given a static scene, we hope to composite the local SDFs for all static objects in the offline stage, to describe the geometry of the given scene. But the local SDFs are defined in different frames.

According to {\bf Assumption \ref{Assum-static-map-be-known}}, the pose $\left[{}^W{\vec p}_{\mathcal{S}_i}, {}^W{\mat R}_{\mathcal{S}_i}\right]$ of each static obstacle $\mathcal{S}_i$ is known and remains unchanged during the manipulator's operation. Let's consider a query point ${}^W{\vec p}$ expressed in the world frame, we can obtain its SDF value by:
\begin{align}
    \label{Eq_SDF_composition}
    \begin{split}
        SDF_{\mathcal{S}}\left({}^W{\vec p}\right) = \min\{  & SDF_{\mathcal{S}_0}\left({}^W{\vec p}\right), SDF_{\mathcal{S}_1}\left({}^{\mathcal{S}_1}{\vec p}\right),\\
                                          &\cdots,  SDF_{\mathcal{S}_{M-1}}\left({}^{\mathcal{S}_{M-1}}{\vec p}\right) \},
    \end{split}
\end{align}
where ${}^{\mathcal{S}_i}{\vec p} = {}^W{\mat R}^\top_{\mathcal{S}_i}\left({}^W{\vec p} - {}^W{\vec p}_{\mathcal{S}_i}\right)$ represents the transformation of a point ${\vec p}$ from $\{W\}$ to $\{\mathcal{S}_i\}$, the $\min\left(\cdots\right)$ means that we would choose the shortest of all distances to each object as the signed distance for the given point.

For a given point ${}^W{\vec p}$, it is possible that this point is out the scope of $E_{\mathcal{S}_i}$. When this happens, it means that this point is far away from the static object $\mathcal{S}_i$ and there is no ``collision-risk" to $\mathcal{S}_i$ at this point. Therefore, we can directly remove $SDF_{\mathcal{S}_i}\left({}^{\mathcal{S}_i}{\vec p}\right)$ from Eq. (\ref{Eq_SDF_composition}).

\subsection{ 4.3 Construction of $Gradient_{\mathcal{L}_i}\left({\vec p}\right)$}  \label{sec:4.3}

On the basis of $SDF_{\mathcal{L}_i}\left({\vec p}\right)$, we can calculate the $Gradient_{\mathcal{L}_i}\left({\vec p}\right)$, i.e., the gradient information at each point in the workspace. The gradient $Gradient_{\mathcal{L}_i}({\vec v})=[g_x,g_y,g_z]^\top$ is given by:
\begin{align}
    \label{Eq_gradient}
    \begin{split}
        g_x &= \frac{SDF_{\mathcal{L}_i}(v_{i+1},v_j,v_k)-SDF_{\mathcal{L}_i}(v_{i-1},v_j,v_k)}{2\delta_{\mathcal{L}_i}}\;,\\
        g_y &= \frac{SDF_{\mathcal{L}_i}(v_i,v_{j+1},v_k)-SDF_{\mathcal{L}_i}(v_i,v_{j-1},v_k)}{2\delta_{\mathcal{L}_i}}\;,\\
        g_z &= \frac{SDF_{\mathcal{L}_i}(v_i,v_j,v_{k+1})-SDF_{\mathcal{L}_i}(v_i,v_j,v_{k-1})}{2\delta_{\mathcal{L}_i}}\; ,
    \end{split}
\end{align}
where ${\vec v} = \left[v_i, v_j, v_k\right]^\top$ is the coordinate of a voxel.

\subsection{4.4 Construction of $SDF_{EE}\left({\vec p}\right)$ and $SDF_{{EE}_i}\left({\vec p}\right)$}  \label{sec:4.4}

\begin{figure}[h]
    \centering
    \subfigure[]{\includegraphics[width = 0.56\hsize]{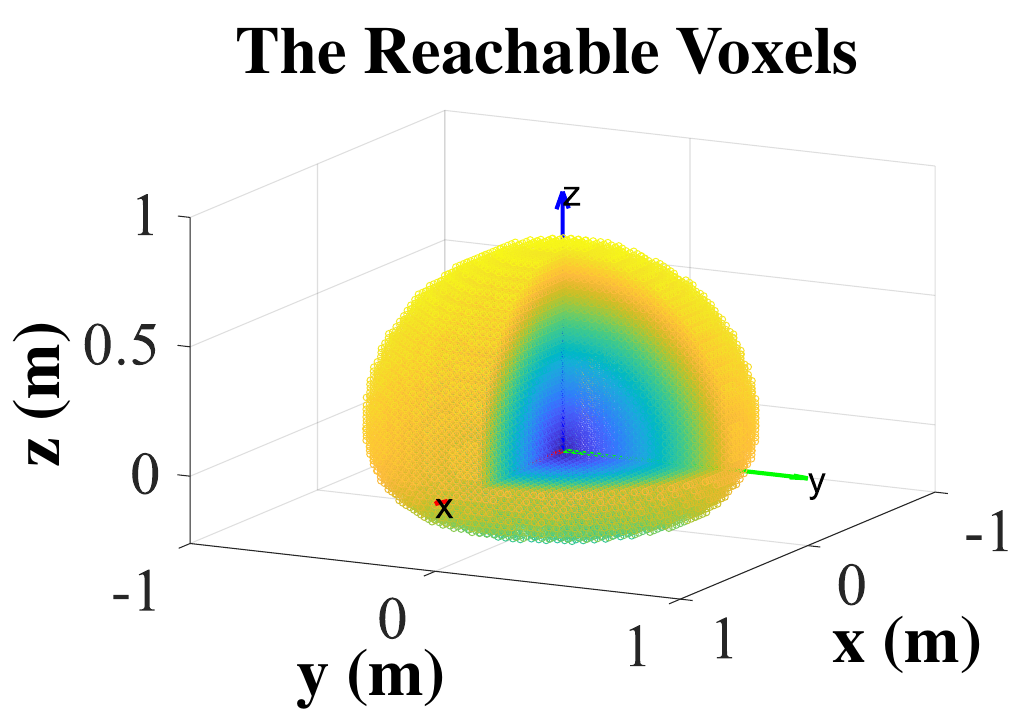}\label{reachable_voxels}}
    \hspace{0.01\hsize}
    \subfigure[]{\includegraphics[width = 0.41\hsize]{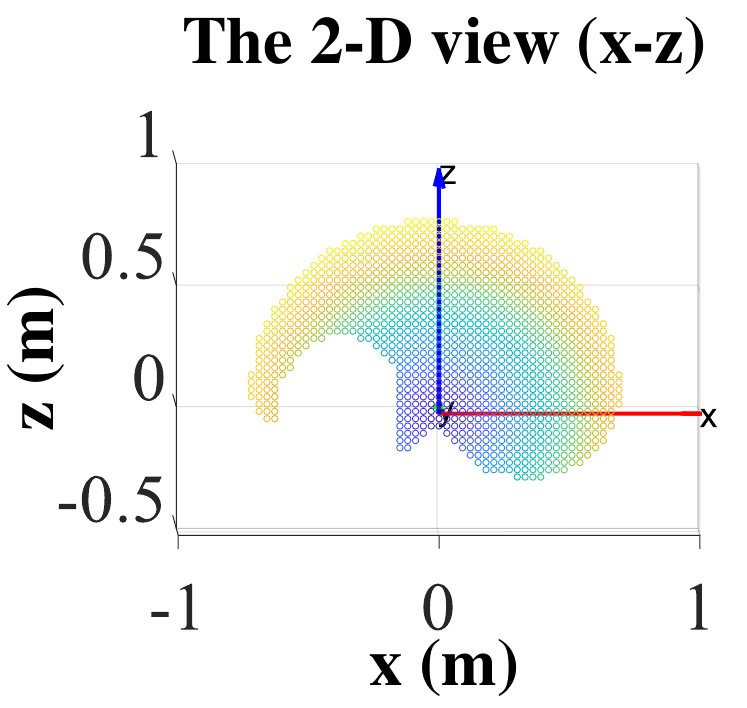}\label{reachable_voxels_2D}}
    \caption{(a) represents the visualization of the reachable voxels of the end-effector of manipulator Unitree Z1 Pro. (b) is a slice of the reachable voxels. }
    \label{the_reachable_voxles}
\end{figure}

Except for the local SDFs that represent shapes of the static objects and the links, we also define the $SDF_{EE}\left({\vec p}\right)$ and $SDF_{{EE}_i}\left({\vec p}\right)$ to represent the reachability of the manipulator's end-effector and each link's end. These SDFs can be utilized to prune the point clouds, because the unreachable points have no ``collision-risk" with the manipulator. However, different to tangible static objects or the links, how to determine the reachable workspace for the manipulator is of the utmost importance.

In order to construct $SDF_{EE}\left({\vec p}\right)$ and $SDF_{{EE}_i}\left({\vec p}\right)$, we first create the enclosure box $E_{\mathcal{M}}$ that encompasses the entire workspace of the manipulator's end-effector based on its kinematic parameters. Denote the upper and lower limits of $E_\mathcal{M}$ in the $X$, $Y$ and $Z$ directions as $X_{max}$, $Y_{max}$, $Z_{max}$, $X_{min}$, $Y_{min}$, $Z_{min}$, respectively. And denote $\delta_{\mathcal{M}}$ as the resolution to discretize $E_\mathcal{M}$. $N_x$, $N_y$, $N_z$ are the number of voxels in the $X$, $Y$ and $Z$ directions, respectively.

\begin{algorithm}
\caption{Identify the reachable and unreachable voxels for a manipulator}\label{algorithm_reachableWS_determination}
\KwIn{$E_{\mathcal{M}}$, $X_{max}$, $X_{max}$, $Z_{max}$, $X_{min}$, $Y_{min}$, $Z_{min}$, $\delta_{\mathcal{M}}$, and the forward kinematics of the manipulator}
\KwOut{$SDF_{EE}$}
$Set\ \  SDF_{EE}({\vec v}) = \overline{d}\ \ \ \  for\ all\ {\vec v} \in E_{\mathcal{M}} $\;

\For{$Every\ {\vec \theta} \in \mathcal{J}$}{
    ${\vec p} = Forwards({\vec \theta}) $\;

    ${\vec v} = Grid({\vec p}) $\;

    \If{$ SDF_{EE}({\vec v}) == \overline{d}$}{
        $Set\ \  SDF_{EE}({\vec v}) = -\overline{d}$\;
    }
}
\end{algorithm}

Without any prior information, we first initialize all voxels in the SDF with a positive value $\overline{d}$ ($\overline{d} > 0$), which is the upper bound for all the possible distances. Then, by traversing the joint space $\mathcal{J}$, we can obtain the reachable position ${\vec p} = \left[p_x,p_y,p_z\right]^\top$ of the manipulator's end-effector in the workspace by using forward kinematics. The obtained position information of the end-effector is rasterized by a grid function $Grid(p)$:
\begin{align}
    \label{Eq_Grid_func}
    \begin{split}
        v_i &= \min(\left\lfloor\frac{p_x-X_{min}}{\delta_{\mathcal{M}}}\right\rfloor +1,N_x)\;,\\
        v_j &= \min(\left\lfloor\frac{p_y-Y_{min}}{\delta_{\mathcal{M}}}\right\rfloor +1,N_y)\;,\\
        v_k &= \min(\left\lfloor\frac{p_z-Z_{min}}{\delta_{\mathcal{M}}}\right\rfloor +1,N_z)\;,
    \end{split}
\end{align}
where $\left\lfloor \cdot \right\rfloor$ represents the floor function. Then the corresponding voxel ${\vec v} = \left[v_i,v_j,v_k\right]^\top$ is assigned a negative value $-\overline{d}$, to indicate that it is reachable by the manipulator's end-effector.

\begin{figure}[ht]
    \centering
    \includegraphics[width = 0.85\hsize]{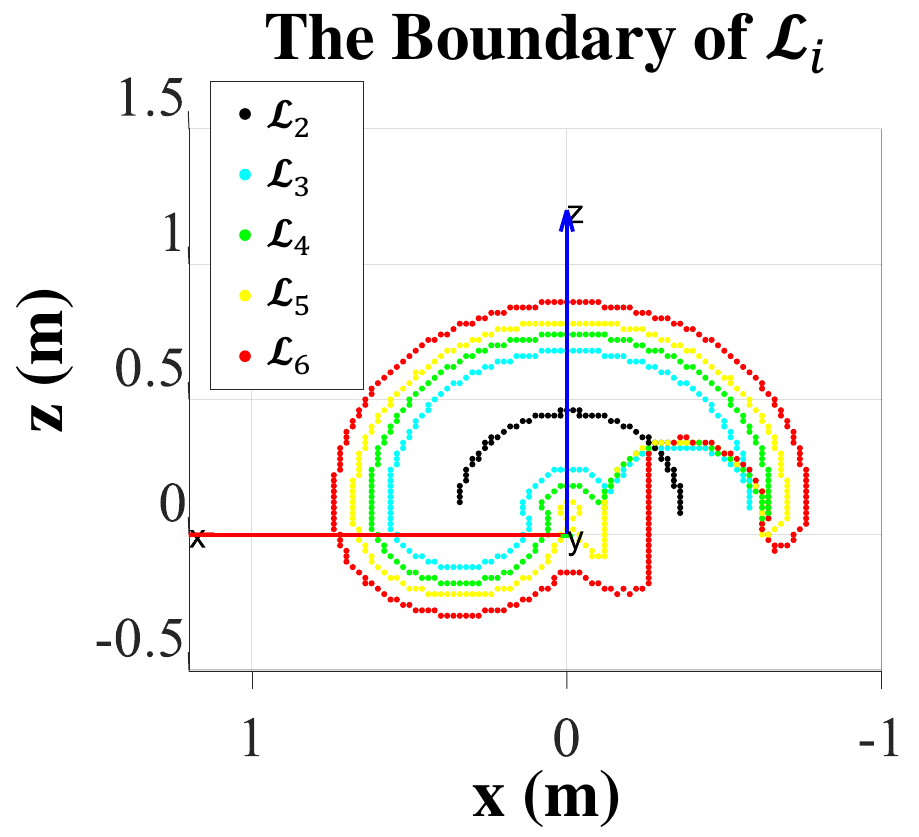}
    \caption{Example of planner boundaries of each link's reachable space. }
    \label{reachable_boundary}
\end{figure}

This process is illustrated in {\bf Algorithm \ref{algorithm_reachableWS_determination}}. Based on {\bf Algorithm \ref{algorithm_reachableWS_determination}}, all the voxels have been classified into two groups: the reachable voxels with negative values and the unreachable voxels with positive values. Fig. \ref{the_reachable_voxles} visualized the reachable voxels of the manipulator Unitree Z1 Pro. Then similar to the calculation of $SDF_{\mathcal{S}_i}\left({\vec p}\right)$ and $SDF_{\mathcal{L}_i}\left({\vec p}\right)$, we can construct the $SDF_{EE}\left({\vec p}\right)$ by \textbf{Algorithm \ref{algorithm_ConstructionOfSiLi}}.

The method described above is applicable not only for computing the $SDF_{EE}\left({\vec p}\right)$, but also for computing the $SDF_{{EE}_i}\left({\vec p}\right)$. To calculate the $SDF_{{EE}_i}\left({\vec p}\right)$, we can ignore the links after $\mathcal{L}_i$ and regard the manipulator as a new manipulator with $i$-DoFs. And the boundaries of each link's reachable space are also visualized in Fig. \ref{reachable_boundary}.

Please note that the $SDF_{\mathcal{L}_i}\left({\vec p}\right)$, $Gradient_{\mathcal{L}_i}\left({\vec p}\right)$, $SDF_{EE}\left({\vec p}\right)$, and $SDF_{EE_i}\left({\vec p}\right)$ are independent with the manipulator's joint states and its working environment. Therefore, for a specific manipulator, the above local SDFs only need to be pre-computed and stored once.

Based on the $SDF_{\mathcal{S}}\left({\vec p}\right)$ computed in Sect. 4.2 and $SDF_{{EE}_i}\left({\vec p}\right)$ computed in Sect. 4.4, we can extract the static obstacles which intrude into the reachable space of each link's end in the offline stage with the following equation:
\begin{equation}
    \hat{\mathcal{S}}_{\mathcal{L}_i} = \left\{ \vec{v} \mid  {\vec v} \in E_\mathcal{S},\, SDF_S\left( \vec{v} \right) < 0, \,SDF_{EE_i}\left( \vec{v} \right)<0 \right\},
\end{equation}
where $\hat{\mathcal{S}}_{\mathcal{L}_i}$ represents the set of all static obstacles which intrude into the reachable space of $\mathcal{L}_i$. The voxels in $\hat{\mathcal{S}}_{\mathcal{L}_i}$ can serve as query points for $SDF_{\mathcal{L}_i}\left({\vec p}\right)$ to generate escape velocity for static objects.

\section{5. Online Unconditioned Reflex Mechanism}

\begin{figure*}[ht]
    \centering
    \includegraphics[scale = 0.52]{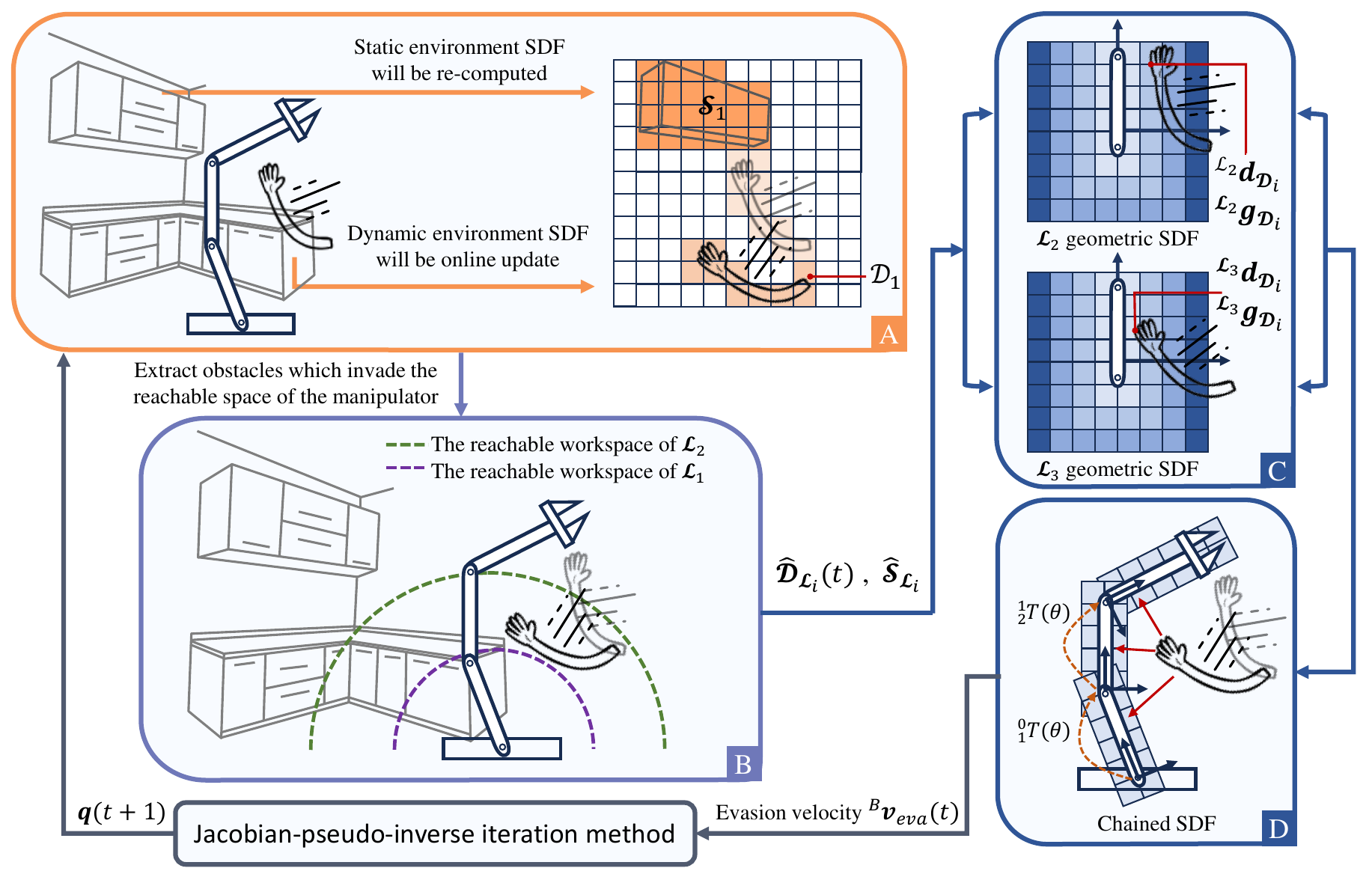}
    \caption{Figure A illustrates the problem scenario in our setup, characterized by an unstructured environment containing a mixture of dynamic and static obstacles, and the SDF that describes the geometric occupancy information of the working environment. Static obstacles are precomputed and stored during the offline phase, while dynamic obstacles are iteratively updated online. Figure B demonstrates the obstacles after extraction by the SDF which describes the reachable space at the end of each link. Figure C shows the SDF which describes each link's geometric information and how we use the SDF to quickly obtain the closest distance between obstacles and link's surface. Figure D illustrates how these link geometric SDFs are chained together through transformation matrices to generate the escape velocity ${{}^B{\vec v}_{eva} \left(t\right)}$. Then the joint actions ${\vec q}\left(t+1\right)$ is generated by the Jacobian-pseudo-inverse iteration method. }
    \label{Fig_whole_process}
\end{figure*}

In this section, we are concern with the point-to-point trajectory planning problem of the manipulator in dynamic environments. More spcifically, we design an unconditioned reflex mechanism such that the manipulator can generate the escape velocity in real-time and take actions to avoid obstacles immediately, while moving towards the target pose. As illustrated in Fig. \ref{Fig_whole_process}, our unconditioned reflex mechanism is divided into two steps.

Step 1: Utilize the chained local SDFs to quickly generate the desired escape velocity ${\vec \upsilon}\left(t\right)$.

Step 2: Transform the escape velocity ${\vec \upsilon}\left(t\right)$ into the desired joint action ${ {\vec q}}\left(t+1\right)$.

\begin{figure}[ht]
    \centering
    \includegraphics[width = 0.9\hsize]{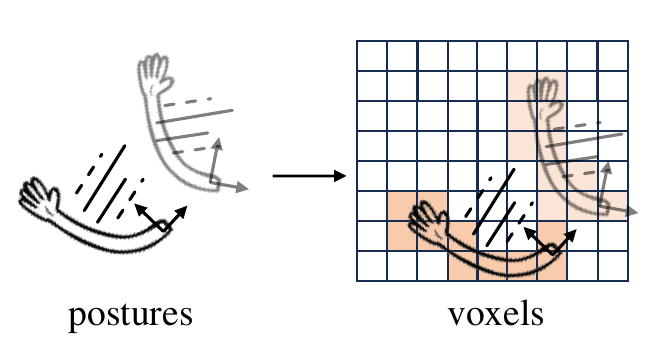}
    \caption{Illustration of a 2D example with an arm and voxels: A dynamic obstacle with an irregular but known shape updates $\mathcal{D}\left( t \right)$ after acquiring its pose.}
    \label{dynamic_obs_update}
\end{figure}

\subsection{5.1 The Query Points Generated from Static and Dynamic Obstacles }

In the end of Sect. 4.3, the query points for static obstacles have been obtained as $\hat{\mathcal{S}}_{\mathcal{L}_i}$ from the composited SDF $SDF_{\mathcal{S}}\left(\vec p\right)$. For the dynamic obstacles, the query points need to be updated online.

For dynamic obstacles whose shapes are pre-known, the occupied points of such an object can be extracted from the local SDF fixed on it. As illustrated in Fig. \ref{dynamic_obs_update}, with knowing the varying poses of this object, the point clouds $\mathcal{D}\left(t\right)$ can be updated in real-time. For unpredictable dynamic obstacles, $\mathcal{D}\left(t\right)$ can be perceived in real time by sensors like RGB-D camera or Lidar.

However, it is not necessary to check all the points in $\mathcal{D}\left(t\right)$. Based on $SDF_{EE_i}\left({\vec p}\right)$, we can prune the points that are unreachable for ${\mathcal{L}_i}$ to accelerate the reflex process. Based on the updated $\mathcal{D}\left(t\right)$, we also extract the portion of dynamic obstacles which intrude into the reachable workspace of $\mathcal{L}_i$ using the following form:
\begin{equation}
    \hat{\mathcal{D}}_{\mathcal{L}_i}\left(t\right) = \left\{ \vec{v} \mid \vec{v} \in \mathcal{D}\left( t \right), SDF_{EE_i}\left( \vec{v} \right)<0 \right\},
\end{equation}
where $\hat{\mathcal{D}}_{\mathcal{L}_i}\left(t\right)$ represents the query points that have collision risk with the link $\mathcal{L}_i$.

The escape velocity for link $\mathcal{L}_i$ can be generated with the $\hat{\mathcal{S}}_{\mathcal{L}_i}$ and $\hat{\mathcal{D}}_{\mathcal{L}_i}\left(t\right)$ being inputs.

\subsection{5.2 Real-time Generation of Desired Escape Velocity Using the Chained local SDFs }

During the offline stage, we pre-compute and store the local SDFs that describes the geometric information of the manipulator and its working environments. However, these local SDFs are defined in different frames. In this subsection, we would like to illustrate how to chain the local SDFs to generate the desired escape velocity for the manipulator.

For efficiency, we have already extracted the set of obstacle voxels $\hat{\mathcal{S}}_{\mathcal{L}_i}$ and $\hat{\mathcal{D}}_{\mathcal{L}_i}\left(t\right)$ which intrude into the reachable space of each link. We assume that the $k^{th}$ obstacle voxel in these sets is $^W{\vec p}_{obs-k}$, which is expressed in the world frame $\{W\}$. However, the local SDFs $SDF_{\mathcal{L}_i}$ and $Gradient_{\mathcal{L}_i}$ are both defined in each link frame $\{\mathcal{L}_i\}$. These local SDFs can be chained according to the forward kinematics of the manipulator. In order to utilize $SDF_{\mathcal{L}_i}$ and $Gradient_{\mathcal{L}_i}$, we need to transform $^W{\vec p}_{obs-k}$ from the world frame $\{W\}$ to the link frame ${\mathcal{L}_i}$:
\begin{align}
    \label{Eq_queryPointsTransformation}
    \begin{split}
      ^{\mathcal{L}_i}{\vec p}_{obs-k}= \, ^{\mathcal{L}_i}_{B} T\left( \vec q\left(t\right) \right) \cdot \, ^{B}_W T \cdot \, ^W{\vec p}_{obs-k}\:,
    \end{split}
\end{align}
where ${\vec q}\left(t\right)$ is the current joint states and $^{B}_W T$ is the transformation from the world frame to the base frame of the manipulator. Without loss of generality, the world frame $\{W\}$ is aligned with the base frame $\{B\}$, i.e., $^{B}_W T = I_4$. Please note that $^{\mathcal{L}_i}_{B} T\left( \vec q\left(t\right) \right) = \, ^{B}_{\mathcal{L}_i} T^{-1}\left( \vec q\left(t\right) \right)$.

The effect of coordinate transformation is shown in Fig.\ref{Fig_whole_process}(C).
Since the gradient information and the distance information to the surface of $\mathcal{L}_i$ at point $^{\mathcal{L}_i}{\vec p}_{obs-k}$ have already been precomputed and stored as the $SDF_{\mathcal{L}_i}\left(^{\mathcal{L}_i}{\vec p}_{obs-k}\right)$ and $Gradient_{\mathcal{L}_i}\left(^{\mathcal{L}_i}{\vec p}_{obs-k}\right)$ during the offline stage, we can obtain the distance and gradient directly by an extremely fast lookup:
\begin{equation}
  d_{obs-k}=SDF_{\mathcal{L}_i}\left( ^{\mathcal{L}_i}{\vec p}_{obs-k}\right)\;,
\end{equation}
\begin{equation}
  ^{\mathcal{L}_i}{\vec g}_{obs-k}=Gradient_{\mathcal{L}_i}\left(^{\mathcal{L}_i}{\vec p}_{obs-k}\right)\;.
\end{equation}

The expected escape velocity is obtained through a linear mapping of the gradient information:
\begin{equation}
  ^{\mathcal{L}_i}{\vec v}_{eva}\left(t\right)= \lambda ^{\mathcal{L}_i}{\vec g}_{obs-k} \;.
  \label{Li_v_eva}
\end{equation}

\begin{figure}[ht]
    \centering
    \includegraphics[scale = 0.35]{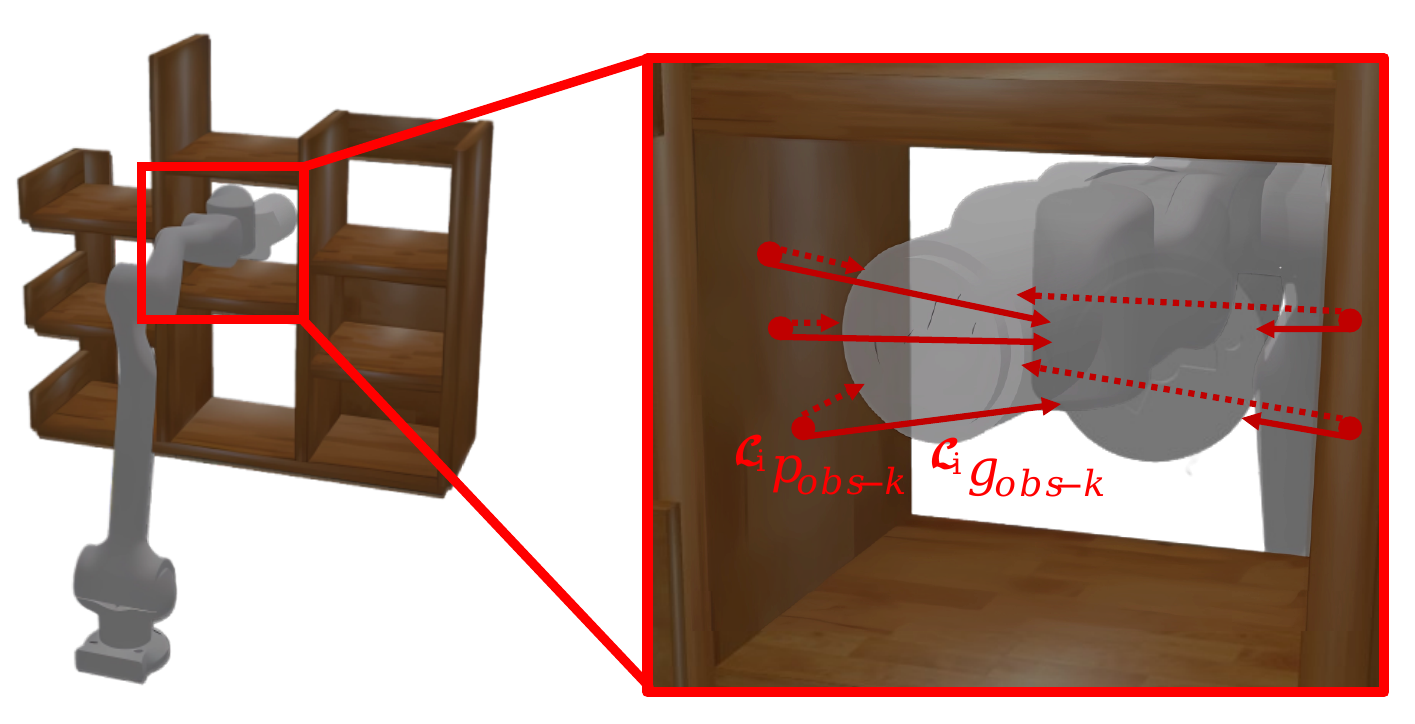}
    \caption{When the manipulator moves deeper into the cabinet, it is essential to ensure that the velocity vector remains continuous to avoid end jitter, which needs to consider all obstacle points in a distance $r_{eva}$.}
    \label{surronding_obstacle}
\end{figure}

However, when complex obstacles surround the manipulator, as shown in Fig.~\ref{surronding_obstacle}, relying solely on the distance to the nearest obstacle point can result in discontinuities or oscillations in the arm's velocity during movement. This approach also diminishes the advantages of the SDF method, which is known for its high accuracy and fast online querying. To prevent velocity discontinuities, we sum all obstacle points within a distance $r_{eva}$ and multiply each by its respective distance weight to generate a continuous desired escape velocity. Finally, an coefficient based on the distance to the nearest obstacle point is applied to ensure the effectiveness of the avoidance. Therefore, Eq.~\eqref{Li_v_eva} is modified as follows:
\begin{align}
    \label{modified_Li_v_eva}
    \begin{split}
      ^{\mathcal{L}_i}{\vec v}_{eva}\left(t\right) &= \gamma  \sum_{k=1} \lambda _k Gradient_{\mathcal{L}_i}\left(^{\mathcal{L}_i}{\vec p}_{obs-k}\right)\;, \\
      ^{\mathcal{L}_i}{\vec p}_{obs-k} &= \left\{^{\mathcal{L}_i}{\vec p}_{obs} \vert SDF_{\mathcal{L}_i}\left( ^{\mathcal{L}_i}{\vec p}_{obs-k}\right) < r_{eva} \right\}\;, \\
      \lambda _k &= \frac{r_{eva} - SDF_{\mathcal{L}_i}\left(^{\mathcal{L}_i}{\vec p}_{obs-k}\right)}{\Sigma\left(r_{eva} - SDF_{\mathcal{L}_i}\left( ^{\mathcal{L}_i}{\vec p}_{obs-k}\right) \right)}\;,  \\
      \gamma &= \frac{r_{eva} - \min_{^{\mathcal{L}_i}{\vec p}_{obs}} SDF_{\mathcal{L}_i}\left(^{\mathcal{L}_i}{\vec p}_{obs-k}\right)}{r_{eva}}\;.
  \end{split}
\end{align}

Due to the fact that the escape velocity should be relative to the base frame, the transformation is needed as follow:
\begin{equation}
  ^B{\vec v}_{eva}\left(t\right) = ^B_{\mathcal{L}_i}T \left({\vec q}\left(t\right)\right)   \cdot \; ^{\mathcal{L}_i}{\vec v}_{eva} \left(t\right)\;.
\end{equation}

\subsection{5.3 Calculation of the Geometrical Jacobian matrix}

With the desired escape velocity $^B{\vec v}_{eva}\left(t\right)$, it is necessary to calculate the Jacobian matrix to generate obstacle avoidance actions.

The Jacobian matrix under the manipulator operation task can usually be described by the following equation:
\begin{equation}
  \begin{bmatrix}
   ^B{\vec \upsilon}\left(t\right) \\
   ^B{\vec \omega}\left(t\right)
  \end{bmatrix}
  =
  \begin{bmatrix}
    J_p\left( {\vec q}\right) \\
    J_r\left( {\vec q}\right)
  \end{bmatrix}
  \dot{ {\vec q}} \left(t\right)\;,
\end{equation}
where $J_p\left( {\vec q}\right)$ and $J_r\left( {\vec q}\right)$ are the position and orientation Jacobian matrices at configuration $q$, $^B{\vec \upsilon}\left(t\right)$ and $^B{\vec \omega}\left(t\right)$, which are expressed in the manipulator's base frame, are the linear and angular velocity at time $t$. For simplicity, we uniformly use $J_p$ and $J_r$ to denote the position and orientation Jacobian matrices in a given joint configuration.

The traditional method that uses the analytic Jacobian matrix has singularity issues, owing to the representational singularity problem of the representation methods for orientation. Therefore, we would like to derive the geometric Jacobian matrix for the convenience of generating actions. Since the Jacobian matrix maps configuration space to task space, we independently solve the mapping matrix $J_p$ for the linear part and the mapping matrix $J_r$ for the orientation part. We assume that $J_{pi}$ and $J_{ri}$ represent the influence of the $i$-th joint on the position and orientation of the end-effector, respectively. Assuming that the position and orientation of the end-effector in the manipulator base frame under the current joint configuration are ${}^B{\vec p}\left( t \right)$ and ${}^B{\mat R}\left( t \right)$, respectively. ${\vec s}_i\left({\vec q}\left(t\right)\right)$ is an unit vector along the rotation axis direction of the joint angle, and ${\vec p}_i\left({\vec q}\left(t\right)\right)$ is an arbitrary point on the joint rotation axis. ${\vec s}_i\left({\vec q}\left(t\right)\right)$ and ${\vec p}_i\left({\vec q}\left(t\right)\right)$ can be iteratively updated in real-time according to the joint states ${\vec q}\left(t\right)$. Then $J_{pi}$ and $J_{ri}$ for a revolute joint can be calculated using the following formulas:
\begin{equation}
    J_{pi}={\vec s}_i\left({\vec q}\left(t\right)\right) \times ({}^B{\vec p}\left( t \right) - {\vec p}_i\left({\vec q}\left(t\right)\right) )\;,
\end{equation}
\begin{equation}
    J_{ri}={}^B{\mat R}^\top\left( t \right) {\vec s}_i\left({\vec q}\left(t\right)\right)\;.
\end{equation}

If the $i$-th joint is a prismatic joint, the $J_{pi}$ and $J_{ri}$ can be given by:
\begin{equation}
    J_{pi}={\vec s}_i\left({\vec q}\left(t\right)\right)\;,
\end{equation}
\begin{equation}
    J_{ri}= {\vec 0}_{3\times 1}\;.
\end{equation}

Hence we have:
\begin{equation}
    J_p=[J_{p1},J_{p2}, \cdots,J_{pn}]\;,
\end{equation}
\begin{equation}
    J_r=[J_{r1},J_{r2}, \cdots,J_{rn}]\;.
\end{equation}

\subsection{5.4 Unconditioned Action Generation for Obstacle Avoidance  }

In the previous subsection, we obtained the desired velocity $^B{\vec v}_{eva}\left(t\right)$ of the manipulator in the presence of obstacles by using local SDFs. In this subsection, we will introduce our method about how to utilize this desired escape velocity to generate an online action that enables the manipulator to avoid the obstacle while moving toward the target.

We employ the iterative method of the pseudo-inverse of the Jacobin matrix to map the desired trajectory from task space to the configuration space, thereby controlling the manipulator's end-effector to move to the target point.

However, the manipulator cannot move directly to the target configuration due to certain constraints. Therefore, we treat these constraints, caused by environmental obstacles, as a secondary task and use redundant degrees of freedom to address them. These constraints are integrated into the Jacobian pseudo-inverse iterative method, which can be expressed as follows:
\begin{equation}
  \dot{\vec{q}}=J^+\dot{\vec{x}}+(I-J^+J)\dot{{\vec \varphi}}\;,
  \label{Jaco_with_constraint}
\end{equation}  
here the parameters, i.e., the time $t$ or the configuration ${\vec q}\left(t\right)$, are neglected in the expression of $J\left({\vec q}\left(t\right)\right)$ and $\dot{\vec q}\left(t\right)$, for simplicity. While $\dot{{\vec \varphi}}$ is an arbitrary velocity depending on secondary task. The $(I-J^+J)$ part is used to project $\dot{{\vec \varphi}}$ into the null space of $J$, such that the secondary task won't interfere the performance of the primary task.

In this paper, we would independently conduct obstacle avoidance tasks for each link. When the $i$-th link encounters an obstacle point at a distance less than the threshold value, the velocity of the manipulator in the configuration space can be described by two equations:
\begin{equation}
\dot{\vec{q}}=J_{EE}^+\dot{\vec{x}}_{EE}\;,
\label{ee_jaco_map}
\end{equation}
\begin{equation}
\dot{\vec{q}}=J_{i}^+\dot{\vec{x}}_{\mathcal{L}_i}\;.
\label{link_jaco_map}
\end{equation}
Among the two equations, Eq.~\eqref{ee_jaco_map} represents the desired joint velocity generated by the end-effector while Eq.~\eqref{link_jaco_map} are for the different links. $\dot{{\vec x}}_{ \mathcal{L}_i} $ denotes the desired evasive velocity to avoid the link $\mathcal{L}_i$. $J_i$ is the relevant Jacobian matrix of the $i$-th link, which can be given by:
\begin{equation}
  J_i=[J_{p1},J_{p2},\cdots,J_{pi},0,\cdots,0]\;.
\end{equation}

\begin{figure}[b]
    \centering
    \includegraphics[scale = 0.45]{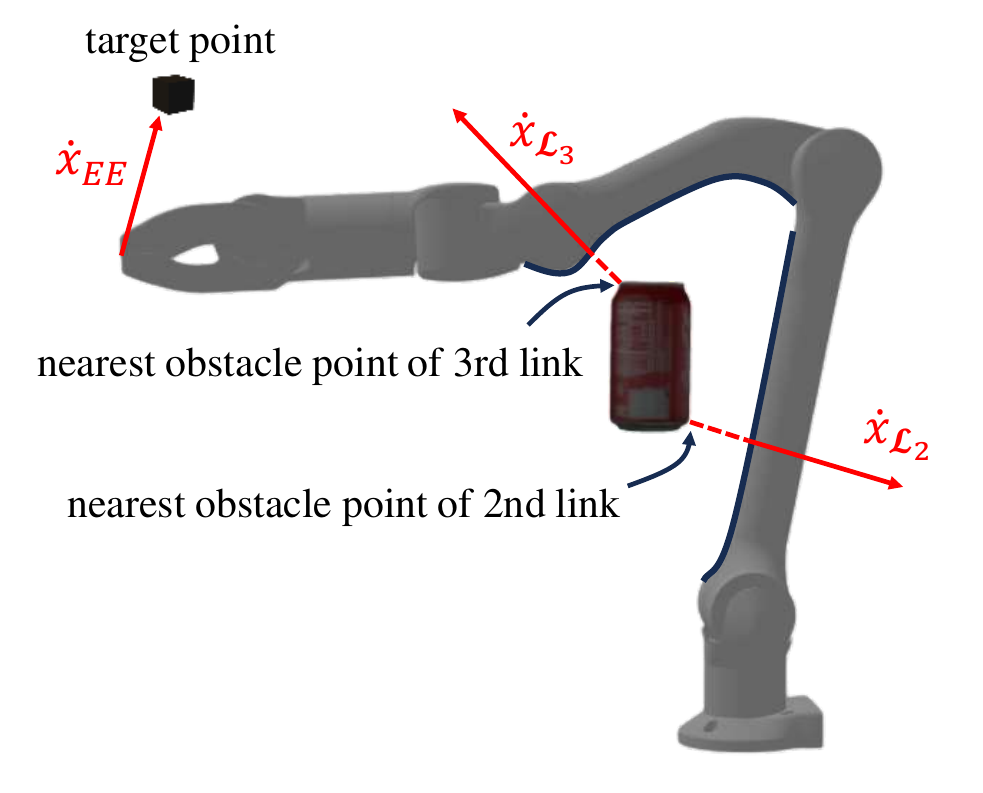}
    \caption{Here, ${\vec x}_{EE}$ represents the primary objective of the manipulator's end movement, while ${\vec x}_{\mathcal{L}_i}$ denotes the secondary objective of obstacle avoidance during the movement of each linkage of the manipulator.}
    \label{Fig_PrimaryAndSecondaryObjective}
\end{figure}

\begin{algorithm}
\caption{Unconditioned action generation }\label{obstacle_avoidance_process}
\KwIn{${\vec q}\left(t\right)$, ${\vec p}\left(t\right)$, ${\vec p}_{tar}$, ${\mat R}\left(t\right)$, ${\mat R}_{tar}$}
\While{$\left\|{\vec p}_{tar} - {\vec p}\left(t\right)\right\| > \epsilon_p$ and $\left\|{\mat R}^\top\left(t\right) {\mat R}_{tar}\right\| > \epsilon_r$}{
    \For{$i=N;i>0;i--$}{
        $^B_{\mathcal{L}_i}T=Forward({\vec q}\left(t\right))$\;
        \eIf{  $i$-th joint is a revolute joint }{
            $J_{pi}={\vec s}_i \times \left({\vec p}\left(t\right)-{\vec p}_i\left(t\right)\right)$\;
            $J_{ri}={\mat R}^{\top}\left(t\right) \cdot {\vec s}_i$\;
        }{
            $ J_{pi}={\vec s}_i$\;
            $ J_{ri}={\vec 0}_{3 \times 1} $\;
        }

    }
    ${\vec \omega}_{e}\left(t\right) = \log({\mat R}^{\top}\left(t\right) \cdot {\mat R}_{tar})$\;
    ${\vec p}_e\left(t\right)= {\vec p}_{tar} -  {\vec p}\left(t\right) $\;
    \If{$\Vert {\vec p}_e\left(t\right) \Vert >3p_{step}$} { \Comment{$p_{step}$ is the length of desired movement of the manipulation per iteration}

        $\Delta {\vec p}_d = \frac{p_{step}}{\Vert {\vec p}_e\left(t\right) \Vert}{\vec p}_e\left(t\right)$\;

        $r_{step} = p_{step}\frac{\Vert {\vec \omega}_{e}\left(t\right) \Vert}{\Vert {\vec p}_e\left(t\right) \Vert}$\;
    }{
        $\Delta {\vec p}_d =\beta {\vec p}_e\left(t\right) $\;
        $r_{step} = \beta \Vert {\vec \omega}_{e}\left(t\right) \Vert$\;
    }
    \eIf{$\Vert {\vec \omega}_{e}\left(t\right) \Vert > \epsilon_r$}{
        $\Delta {\vec \omega}_d ={\vec \omega}_{e}\left(t\right) \frac{r_{step}}{\Vert {\vec \omega}_{e}\left(t\right) \Vert}$\;
    }{
        $\Delta {\vec \omega}_d = 0_{3\times 1}$\;
    }
    \For{$i=1;i<N;i++$}{
        $ ^{\mathcal{L}_i}{\vec v}_{eva} = \gamma  \sum_{k=1} \lambda _k Gradient_{\mathcal{L}_i}(^{\mathcal{L}_i}{\vec p}_{obs-k})$\;
        $^B{\vec v}_{eva} = ^B_{\mathcal{L}_i}T \cdot ^{\mathcal{L}_i}{\vec v}_{eva}$\;
        $J_{EE} = [J_p;J_r]$\;
        $\dot{\vec x}_{EE} = [\Delta {\vec p}_d;\Delta {\vec \omega}_d] $\;
        $\Delta {\vec q}_{main} = J_{EE}^+\dot{\vec x}_{EE}/N $\;
        $\Delta {\vec q}_{eva} = (I-J^+_pJ_p)(J_i(I-J^+_pJ_p))^+(^B{\vec v}_{eva}-J_iJ^+_p \Delta {\vec q}_d)$\;
        \eIf{$^{\mathcal{L}_i}d_{obs} < r_{eva}$}{
            $\Delta {\vec q} = \Delta {\vec q}_{main}+\Delta {\vec q}_{eva}$\;
        }{
            $\Delta {\vec q} = \Delta {\vec q}_{main}$\;
        }
    }
    ${\vec q}\left(t+1\right) = {\vec q}\left(t\right) + \Delta {\vec q} $\;
}
\end{algorithm}

To obtain the $\dot{\vec {q}}$ simultaneously satisfying Eq.~\eqref{ee_jaco_map} and Eq.~\eqref{link_jaco_map}, Eq.~\eqref{link_jaco_map} is combined with Eq.~\eqref{Jaco_with_constraint} to get $\dot{\vec \varphi}$ with the form of \\
\begin{equation}
\dot{{\vec \varphi}}=(J_i(I-J_{EE}^+J_{EE}))^+(\dot{{\vec x}}_{\mathcal{L}_i}-J_iJ_{EE}^+\dot{{\vec x}}_{EE})\;.
\label{constrains}
\end{equation}
Then substitute Eq.~\eqref{constrains} into Eq.~\eqref{Jaco_with_constraint}, the solution of joint velocity can be finally written as :
\begin{equation}
\begin{split}
\dot{\vec q}=J_{EE}^+\dot{\vec x}_{EE}+(I-J_{EE}^+J_{EE})(J_i(I-J_{EE}^+J_{EE}))^+ \\ (\dot{\vec x}_{\mathcal{L}_i}-J_iJ_{EE}^+\dot{\vec x}_{EE})\;.
\end{split}
\end{equation}

We divide the equation into two components: the main task part and the secondary task part. In this paper, the main task involves controlling the manipulator's end-effector, denoted as $J_{EE}=[J_{p};J_{r}]$. The secondary task, which focuses on obstacle avoidance, requires each link of the manipulator to avoid obstacles in the task space. Therefore, $J_p$ can substitute for $J_{EE}$. Additionally, all variables in the above equation are in the manipulator's base frame.As shown in Fig. \ref{Fig_PrimaryAndSecondaryObjective}, for clarity, we define $^B{\vec v}_{eva}\left(t\right)=\dot{\vec x}_{\mathcal{L}_i}$, which is given by Eq.~\eqref{modified_Li_v_eva}, as the obstacle avoidance velocity of each link in the world frame. Thus, the corrected formula is:\\
\begin{equation}
\begin{split}
\dot{\vec q}=J_{EE}^+\dot{\vec x}_{EE}+(I-J_{p}^+J_{p})(J_i(I-J_{p}^+J_{p}))^+ \\ (^B{\vec v}_{eva}\left(t\right)-J_iJ_{p}^+\dot{\vec x}_{p})\;.
\end{split}
\end{equation}

In the point-to-point trajectory planning, the end-effector approaches the target iteratively until Eq.~\eqref{Eq_target} is satisfied. The described obstacle avoidance process method is detailed in \textbf{ Algorithm \ref{obstacle_avoidance_process}}.

Since our approach is based on the real-time escape velocity, we allow the manipulator to leverage redundant degrees of freedom in the null space to avoid obstacles. As a result, the manipulator can both converge toward the target point and avoid obstacles simultaneously. This also ensures the validity of our planned action at each step, which means that our method does not require to wait the finish of the entire path planning, as is the case with sampling-based methods.

\section{6. Simulation and Experimental Evaluations }
\label{Section6_Experiments}

In this chapter, we design two main types of experimental tasks to evaluate the performance of our proposed algorithms. The first type assesses the algorithms in static scenarios, focusing on metrics such as time consumption and path quality while comparing them to the current mainstream sampling-based algorithms. The second type examines the applicability of the algorithms in dynamic scenarios, where both obstacles and target points may vary over time, to evaluate their adaptability to dynamic environments.

In the static scenario experiments, we implement our proposed method alongside the mainstream sampling-based methods on a 6-DoF industrial manipulator within a simulation environment to perform experiments. Additionally, we deploy our method on the Unitree Z1 Pro, a 6-DoF collaborative manipulator, to complete challenging tasks both in simulation and real world.

\begin{figure}[ht]
    \centering
    \subfigure[]{\includegraphics[width = 0.48\hsize]{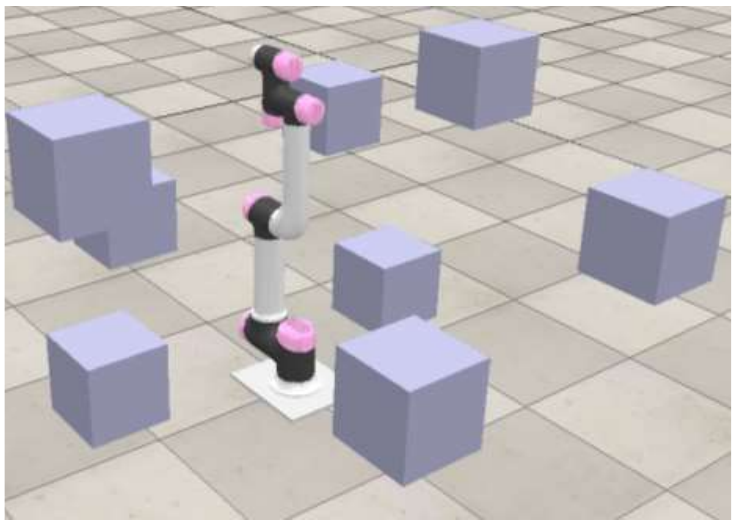}\label{static_scenario_1}}
    \hspace{0.01\hsize}
    \subfigure[]{\includegraphics[width = 0.48\hsize]{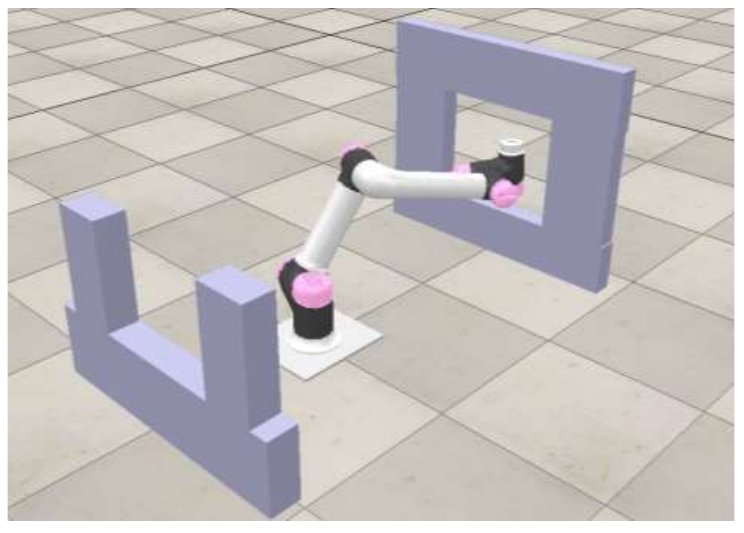}\label{static_scenario_2}}
    \caption{(a) depicts a complex task scenario with eight uniformly sized cubes arranged as obstacles in the space. (b) illustrates a task-oriented scenario where the manipulator must move from one position to the other side of the window without collisions.}
    \label{static_scenario}
\end{figure}

In the experiments involving dynamic scenarios, we conduct both simulation and real-world experiments using the Unitree Z1 Pro to track a dynamic target point while avoiding dynamic obstacles. These experiments are designed to demonstrate the highly-dynamic unconditioned reflex ability of our proposed method.

\begin{figure*}[ht]
    \centering
    \subfigure[ higher is better]{\includegraphics[width = 0.30\hsize]{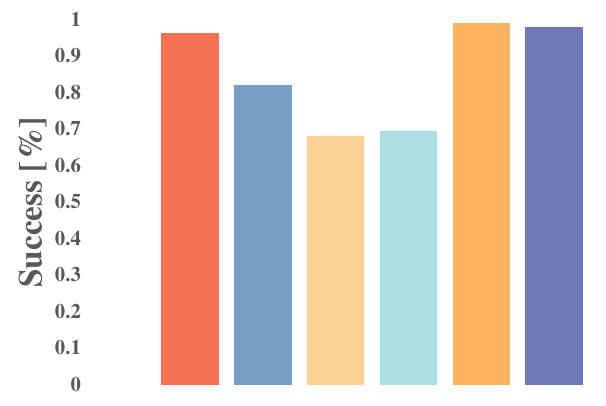}\label{test8_a}}
    \hspace{0.01\hsize}
    \subfigure[ lower is better]{\includegraphics[width = 0.30\hsize]{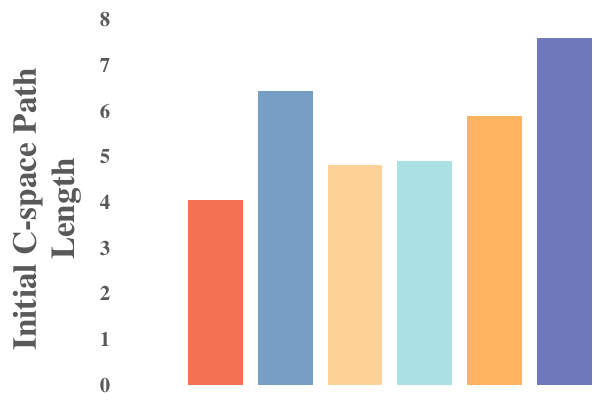}\label{test8_b}}
    \hspace{0.01\hsize}
    \subfigure[ lower is better]{\includegraphics[width = 0.30\hsize]{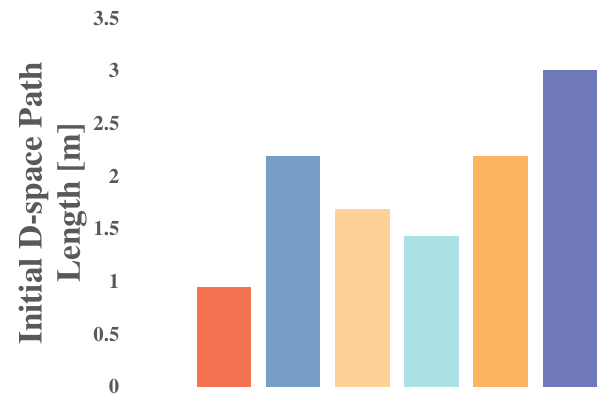}\label{test8_c}}
    \vspace{0.01\hsize}
    \subfigure[ lower is better]{\includegraphics[width = 0.30\hsize]{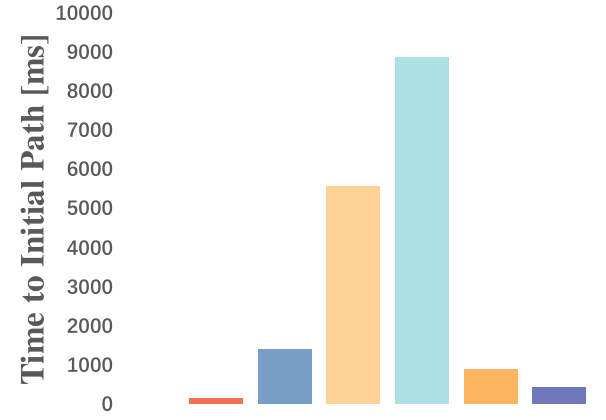}\label{test8_d}}
    \hspace{0.01\hsize}
    \subfigure[ lower is better]{\includegraphics[width = 0.30\hsize]{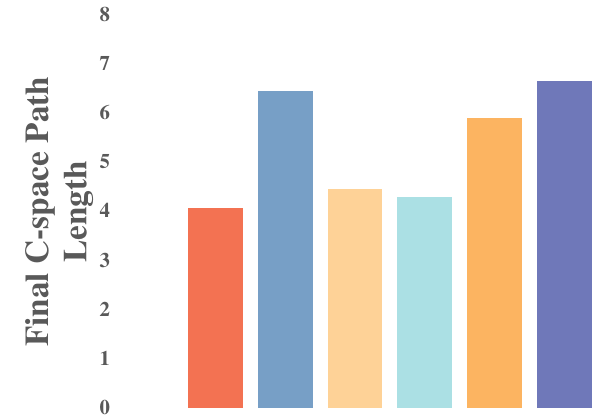}\label{test8_e}}
    \hspace{0.01\hsize}
    \subfigure[ lower is better]{\includegraphics[width = 0.30\hsize]{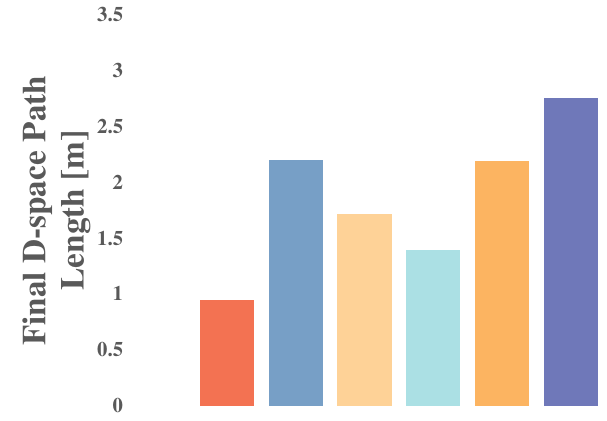}\label{test8_f}}
    \vspace{0.01\hsize}
    \includegraphics[scale = 0.45]{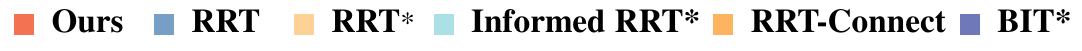}
    \caption{ The experimental results of the manipulator performing in the first type of static scenario, involving 100 sets of 20-second planning for a complex problem. (a) The planning success rate, (b) the average length of the initial path in configuration space, (c) the average length of the initial path in Cartesian space, (d) the average time taken to plan the initial path, (e) the average length of the optimized path in configuration space, and (f) the average length of the optimized path in Cartesian space are presented for each planning algorithm. Our proposed method achieves comparable success rate to the RRT-Connect and BIT*, with only $18.1\%$ and $36.4\%$ of the computational time required By mapping the desired motion direction from Cartesian space to joint space, our method finds solutions with significantly lower path length.}
    \label{test8_result}
\end{figure*}

These experiments are conducted on a laptop equipped with a 2.30 GHz $\times$ 16 Intel Core i7-10875H processor, NVIDIA GetForce RTX 2060 GPU, and 16 GB RAM.

\subsection{6.1 Static Environment Evaluations}
In this subsection, we compare the proposed algorithm with contemporary sampling-based path planning algorithms. Specifically, we evaluate RRT, RRT*, Informed RRT*, RRT-Connect, and BIT*, which represent the state-of-the-art in sampling-based path planning and have robust, high-quality open-source implementations. This experiment assesses the performance of motion planning on a 6-DoF UR-like manipulator, which is manufactured by the Bozhilin Robotic Ltd.

Fig. \ref{static_scenario} presents two types of planning scenarios designed for the experiment. Fig.\ref{static_scenario_1} depicts a complex scenario involving eight cubic obstacles, each with side lengths of 20 cm, positioned randomly within the manipulator's workspace. In this scenario, the manipulator is given randomized initial and target joint configurations. Fig.\ref{static_scenario_2} illustrates a task-oriented scenario where a wall exists within the workspace, featuring a window with side lengths of 40 cm. The manipulator must plan a collision-free trajectory from randomly generated joint configurations to pass its end-effector through the window. All algorithms will be tested simultaneously with the same randomly generated joint configurations to ensure comparability of the experimental results.

The planning time for all planners was limited to 10 seconds, with experiments that failed to plan a trial within 10 seconds being judged as failures. However, for the three algorithms that require path optimization, i.e., the RRT*, the Informed RRT*, and the BIT* we relax the time limit to 20 seconds, considering the time consumption of the optimization process. Both the planned initial paths and the optimized paths within 20 seconds are recorded. All the planners use the Euclidean distance in joint space or Cartesian space as the metric, with the target thresholds of end position deviation being less than 0.06m and the attitude/orientation deviation being less than 0.15rad.

Some key parameters for the SOTA algorithms are given as follows. For exampe, the RRT, the RRT*, and the Informed RRT* planners use a $5\%$ goal-sampling bias. The maximum edge length for the RRT* and the Informed RRT* is set as 0.7rad. The rewiring technique (\cite{2011_RSJ_Asymptotically_optimal_path_planning}) is used in both the RRT* and Informed RRT* methods for path optimization. For all the sampling algorithms based on RRT, the collision detection is performed using the FCL collision detection library. The BIT* algorithm sets the connectivity radius as the twice of the step length and sets 130 samples in each batch. While the step length is set as 0.25rad in configuration space. Our algorithm proposed in this paper sets $r_{eva} = 0.25{\text m}$ as the threshold for enabling the obstacle avoidance process. For fair comparisons, all these parameters have been fine tuned to achieve the best performance, e.g., aiming at reducing the time-consuming process of finding the feasible path.

We randomly generate several complex experimental scenarios of the first type, as shown in Fig. \ref{static_scenario_1}, each with 100 starting and target point configurations using pseudo-randomized seeds. All the sampling-based algorithms are run three times for each setting. However, our algorithm proposed in this paper is deterministic, so each setting is run only once. We recorded the success rate, the average time-consuming, the length of the initial path in joint space, the length of the initial path in Cartesian space {for the end-effector , the length of the optimized path in joint space, and the length of the optimized path in Cartesian space for the end-effector.

\begin{figure*}[ht]
    \centering
    \subfigure[ higher is better]{\includegraphics[width = 0.30\hsize]{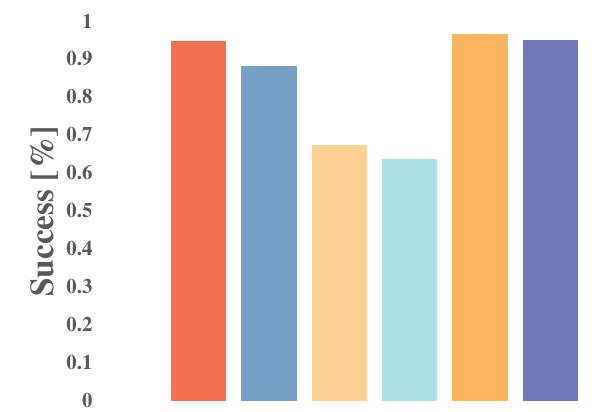}\label{test10_a}}
    \hspace{0.01\hsize}
    \subfigure[ lower is better]{\includegraphics[width = 0.30\hsize]{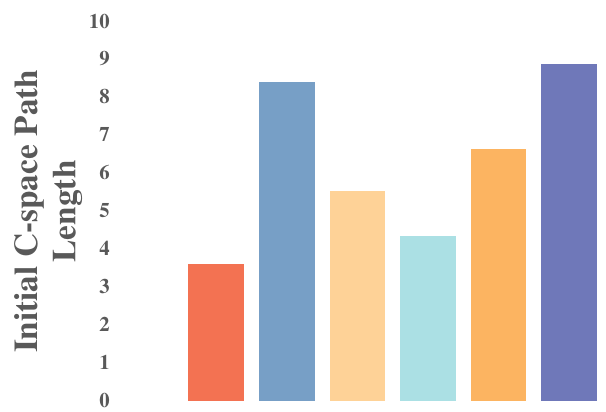}\label{test10_b}}
    \hspace{0.01\hsize}
    \subfigure[ lower is better]{\includegraphics[width = 0.30\hsize]{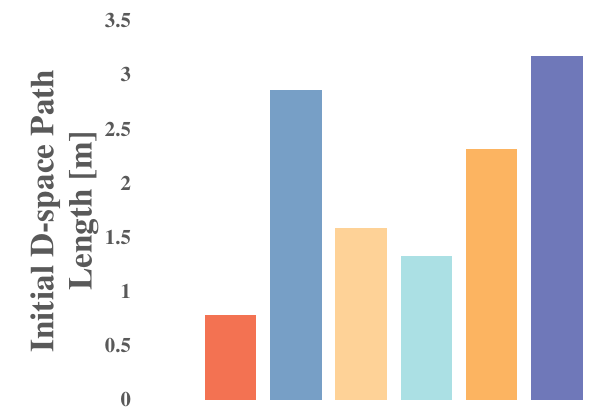}\label{test10_c}}
    \vspace{0.01\hsize}
    \subfigure[ lower is better]{\includegraphics[width = 0.30\hsize]{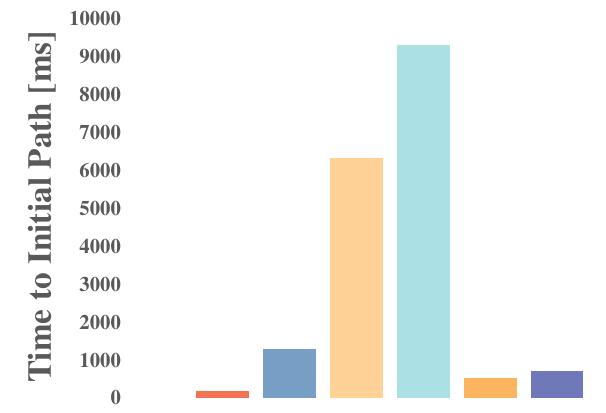}\label{test10_d}}
    \hspace{0.01\hsize}
    \subfigure[ lower is better]{\includegraphics[width = 0.30\hsize]{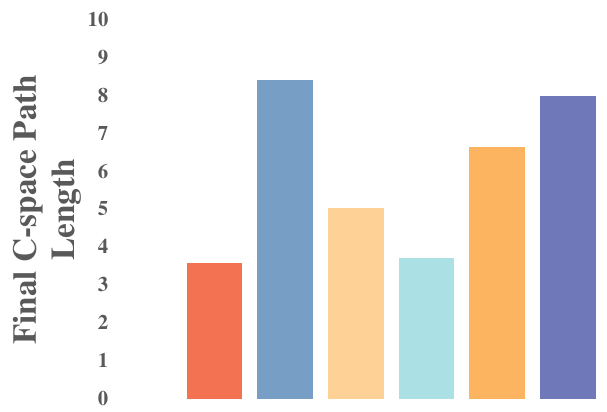}\label{test10_e}}
    \hspace{0.01\hsize}
    \subfigure[ lower is better]{\includegraphics[width = 0.30\hsize]{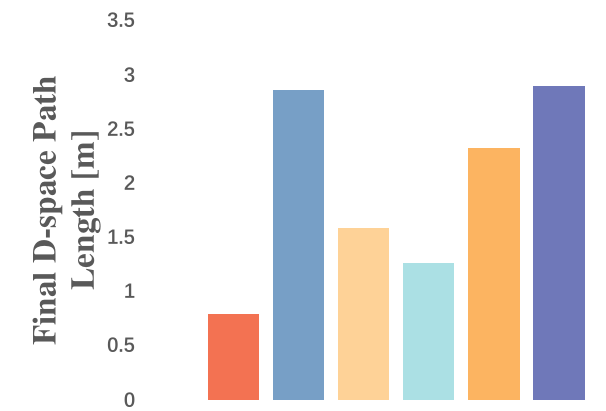}\label{test10_f}}
    \vspace{0.01\hsize}
    \includegraphics[scale = 0.45]{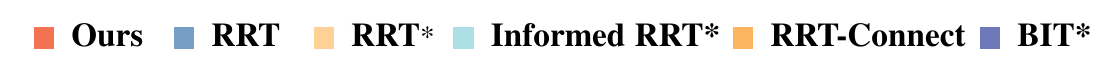}
    \caption{The experimental results of the manipulator in the second type of static scenario, involving 100 sets of 20-second planning for a task-oriented problem. (a) The planning success rate, (b) the average length of the initial path in configuration space, (c) the average length of the initial path in Cartesian space, (d) the average time taken to plan the initial path , (e) the average length of the optimized path in configuration space, and (f) the average length of the optimized path in Cartesian space are presented for each planner. }
    \label{test10_result}
\end{figure*}

The results of this experiment are displayed in Fig.~\ref{test8_result}. In complex experimental scenarios, the proposed method has a success rate close to that of the RRT-Connect and the BIT*, both of which are significantly higher than the success rates of the RRT, the RRT*, and the Informed RRT*. This is because, in high-dimensional spaces, the probability of moving towards the target for the unidirectional-sampling algorithms is greatly reduced, and a high goal-sampling bias makes it difficult to escape from obstacles in complex scenes. Conversely, the RRT-Connect, through bidirectional sampling, and the BIT*, through single-batch multi-point sampling, still perform well in high-dimensional spaces. While our proposed method reduces the effect of dimensionality to a negligible level, since the computational complexity of the lookup at SDFs remains the same.

In terms of time consumption, our proposed algorithm demonstrates significant improvement, averaging about 165ms in the complex experimental scenarios of Fig. \ref{static_scenario_1}. This is a notable enhancement compared to the average time consumption of 453ms for the BIT* algorithm and 910ms for the RRT-Connect. Regarding path lengths, although the RRT*, the Informed RRT*, and the BIT* are asymptotically optimal algorithms and do optimize paths to some extent within a given time, our proposed algorithm still achieves better results in the same time. The average path length of our proposed algorithm in joint space is $91.19\%$ of the RRT* algorithm, $94.40\%$ of the Informed RRT* algorithm, and $61.12\%$ of the BIT* algorithm. Additionally, since our proposed algorithm tracks the desired end-effector trajectory in Cartesian space, its average end-effector trajectory length in Cartesian space is $55.23\%$ of the RRT* algorithm, $68.19\%$ of the Informed RRT* algorithm, and $34.44\%$ of the BIT* algorithm. This indicates that, although the RRT* and the Informed RRT* algorithms have similar performance with ours in joint space, our proposed algorithm shows greater superiority in Cartesian space.

In the experimental scenarios of the second type, as illustrated in Fig. \ref{static_scenario_2}, the manipulator is required to pass through a window. We fix the same target configuration and generate 40 starting configurations using pseudo-random seeds. All the sampling-based algorithms are run three times for the same configuration. The results are shown in Fig. \ref{test10_result}. This experimental scenario is particularly challenging because the target point is surrounded by obstacles, making it difficult for unidirectional-sampling planning algorithms. Consequently, the RRT-Connect algorithm, based on bi-directional sampling planning, has the lowest average time among the sampling-based methods, at about 542ms. However, our proposed algorithm has an average time of 193ms and also achieves a relatively optimal path.

\begin{figure}[ht]
    \centering
    \subfigure[]{\includegraphics[width = 0.85\hsize]{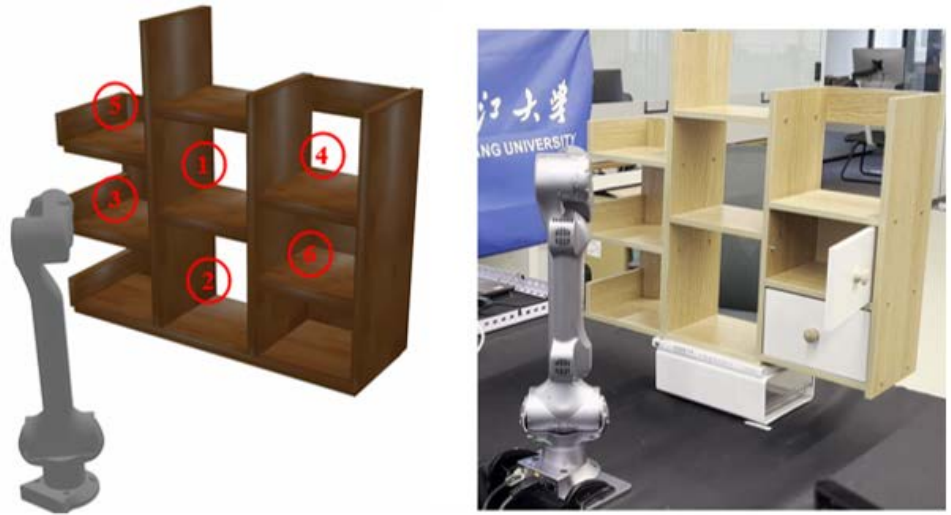}\label{shuli_setup}}
    \vspace{0.001\hsize}
    \subfigure[]{\includegraphics[width = 0.9\hsize]{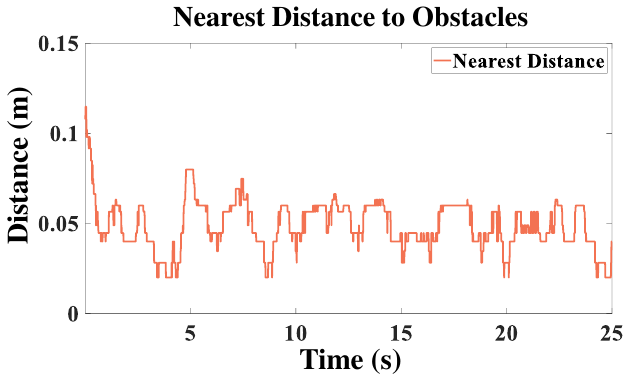}\label{shuli_result}}
    \caption{The scene setup for the super challenging static task is shown in (a), while the experimental result of this task is presented in (b). The order in which the manipulator penetrates deeper into the cabinet is indicated by the numbers in (a). We recorded the distance of the entire manipulator from the nearest obstacle at each moment during the experiment.}
    \label{shuli}
\end{figure}

In addition to the experimental scenarios described above, we also deploy our method on Unitree Z1, a collaborative manipulator with 6-DoF, to execute the task in a super challenging static task scenario. We choose a desktop bookshelf with various cabinet sizes as static obstacles. The manipulator needs to maneuver into different compartments in a specified order. The largest compartment measures 18cm in height and 19cm in width, while the smallest compartment is only 12cm in height and 18cm in width. The simulation scene and results are depicted in Fig.\ref{shuli_setup} and Fig.\ref{shuli_result}, which records the closest distance of the manipulator's links to the bookcase partitions during the simulation.  Additionally, we also conduct a real experimental scenario based on the same setting as in the simulation, with its results available in multimedia.

\subsection{6.2 Dynamic Environment Evaluations }

In this section, we design various dynamic scenarios to evaluate the effectiveness of our proposed algorithm in both dynamic obstacle avoidance and dynamic target tracking. These capabilities are fundamental and crucial for a manipulator working collaboratively with humans or serving people in daily life.

In this experiment, we first evaluate the manipulator's effectiveness in avoiding dynamic obstacles and tracking dynamic target points within a simulated environment. We utilize the Unitree Z1 Pro to test its performance under various conditions: tracking static target points in a dynamic obstacle environment and tracking dynamic target points in a dynamic obstacle environment. The SDFs for obstacles in the simulation environment are pre-calculated and stored.

\subsubsection{6.2.1 Online path planning with dynamic obstacle and static target\\}

This experiment validates the effectiveness of our method in avoiding both dynamic and static obstacles. We created a square cavity with a side length of 35cm on a wall within the simulation environment, placing the target point behind the wall. An obstacle moves at a speed of 5cm/s before and after the target point. The experimental scenario and results are shown in Fig.~\ref{sta_tar_dyn_obs_setup} and Fig.~\ref{sta_tar_dyn_obs_our} and multimedia . When a dynamic obstacle approaches, the manipulator moves backward to evade it, causing an increase in the error of the end effector to target. Once the obstacle recedes, the manipulator converges toward the target point. When the dynamic obstacle appears on the intended path and approaches the manipulator, it retracts to avoid the obstacle.

\begin{figure}[t]
    \centering
    \subfigure[]{\includegraphics[width = 0.75\hsize]{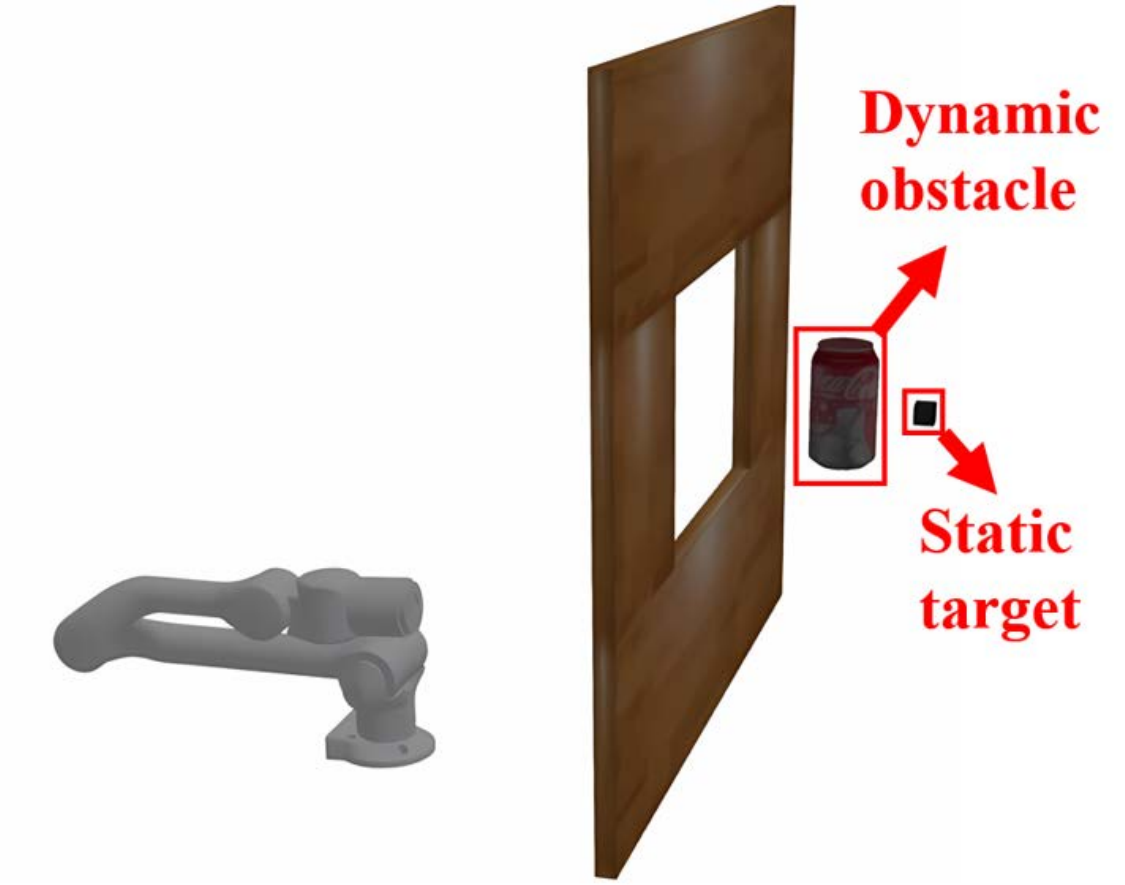}\label{sta_tar_dyn_obs_setup}}
    \vspace{0.001\hsize}
    \subfigure[]{\includegraphics[width = 0.9\hsize]{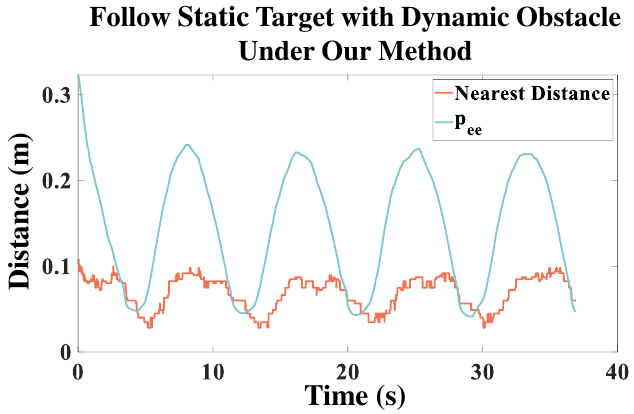}\label{sta_tar_dyn_obs_our}}
    \vspace{0.001\hsize}
    \subfigure[]{\includegraphics[width = 0.9\hsize]{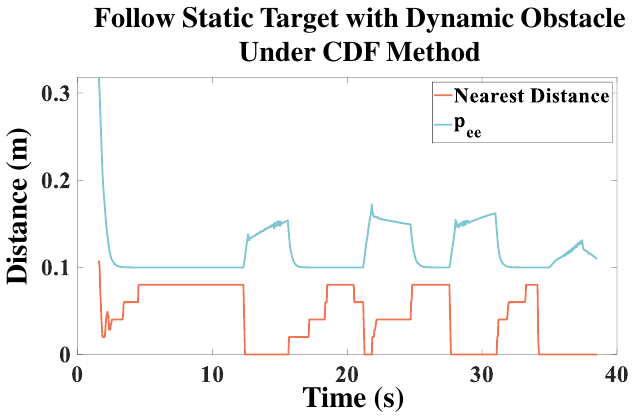}\label{sta_tar_dyn_obs_cdf}}
    \caption{ The simulation setup for the static target and dynamic obstacle task is shown in (a). The experimental result of this task under our proposed approach is shown in (b), while the result under CDF method is shown in (c).}
    \label{cdf_nd_pee}
\end{figure}

Additionally, we implemented another SDF-based state-of-the-art (SOTA) approach, i.e., the Configuration Space Distance Function (CDF) (\cite{2024_arxiv_CDF}), in the Fig.~\ref{sta_tar_dyn_obs_setup} test scenario for comparison. In the CDF, a neural network was trained to describe the distance function of the Unitree Z1 in configuration space, enabling efficient queries for distance and gradient in configuration space. We trained the CDF neural network on our laptops. Due to computing resource limitations, the resolution to discretize the Cartesian space can only be set as 5cm in the Data Generation module, while our algorithm could divide the same size workspace into 1cm voxels. After a 8-hours training, the trained network only achieved an accuracy of $81.26\%$ for the signed distance prediction. The CDF algorithm was subsequently experimentally tested with a quadratic programming controller. The distance between obstacles and manipulator and the distance to the target position were recorded similarly in Fig. \ref{sta_tar_dyn_obs_cdf}. During the experiment, when the manipulator was surrounded by static obstacles in a complex situation, it demonstrated poor obstacle avoidance performance when faced with incoming dynamic obstacle, resulting in multiple collisions. The collisions happen because the trained network is hard to achieve a $100\%$ prediction for the signed distance prediction. This is also an inherent drawback for learning-based approaches. While our approach maintained a distance of at least 2cm from the obstacles throughout the experiment. Additionally, the manipulator planned by CDF exhibited shaking behavior when approaching multiple obstacles.

Meanwhile, We also compared the elapsed time per iteration between our proposed method and the CDF method in this experimental scenario. The results were shown in Fig. \ref{time_consume} and Table~\ref{timecost_label}. The average elapsed time per iteration of our proposed algorithm is 0.818ms, with a root mean square of 1.019ms, while the average elapsed time per iteration of the CDF method is 21.098ms, with a root mean square of 21.207ms. The elapsed time for all iterations is within 5ms, indicating that our algorithm can react to dynamic obstacles and perform obstacle avoidance within 5ms at its slowest in this experiment. This comparison indicate that our proposed method can be more reactive while achieving better performance than SOTA methods.

\begin{figure}[t]
    \centering
    \subfigure[]{\includegraphics[width = 0.95\hsize]{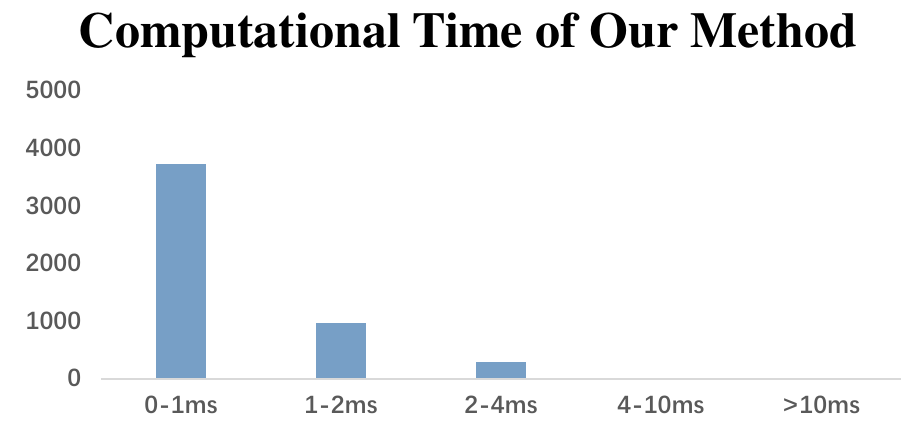}\label{average_timecost}}
    \vspace{0.01\hsize}
    \subfigure[]{\includegraphics[width = 0.95\hsize]{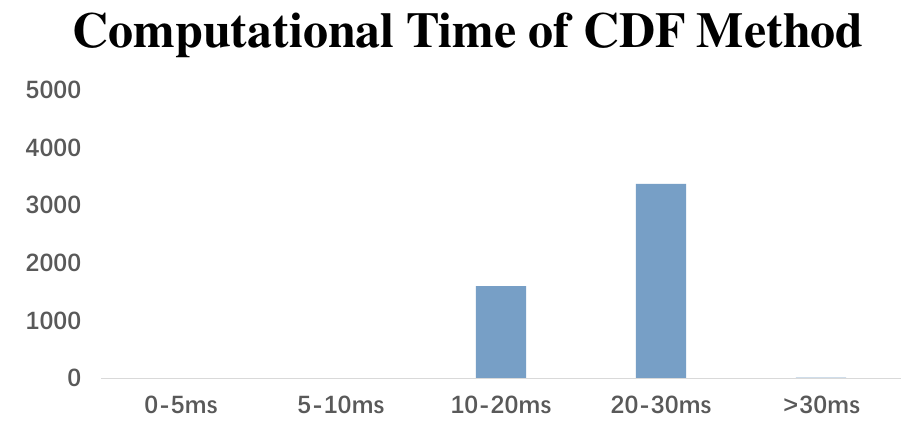}\label{average_timecost_cdf}}
    \caption{Histogram of the time costs in 5000 iterations. (a) Our method. (b) The CDF method. The horizontal axis represents the time intervals, and the vertical axis shows the number of iterations falling within each interval. The time costs of our method is mainly concentrated in 0-1ms, while the time costs of CDF method is mainly concentrated in 20-30ms}
    \label{time_consume}
\end{figure}

\subsubsection{6.2.2 Online path planning with dynamic obstacle and target}

\begin{figure*}[ht]
    \centering
    \subfigure[]{\includegraphics[width = 0.3\hsize]{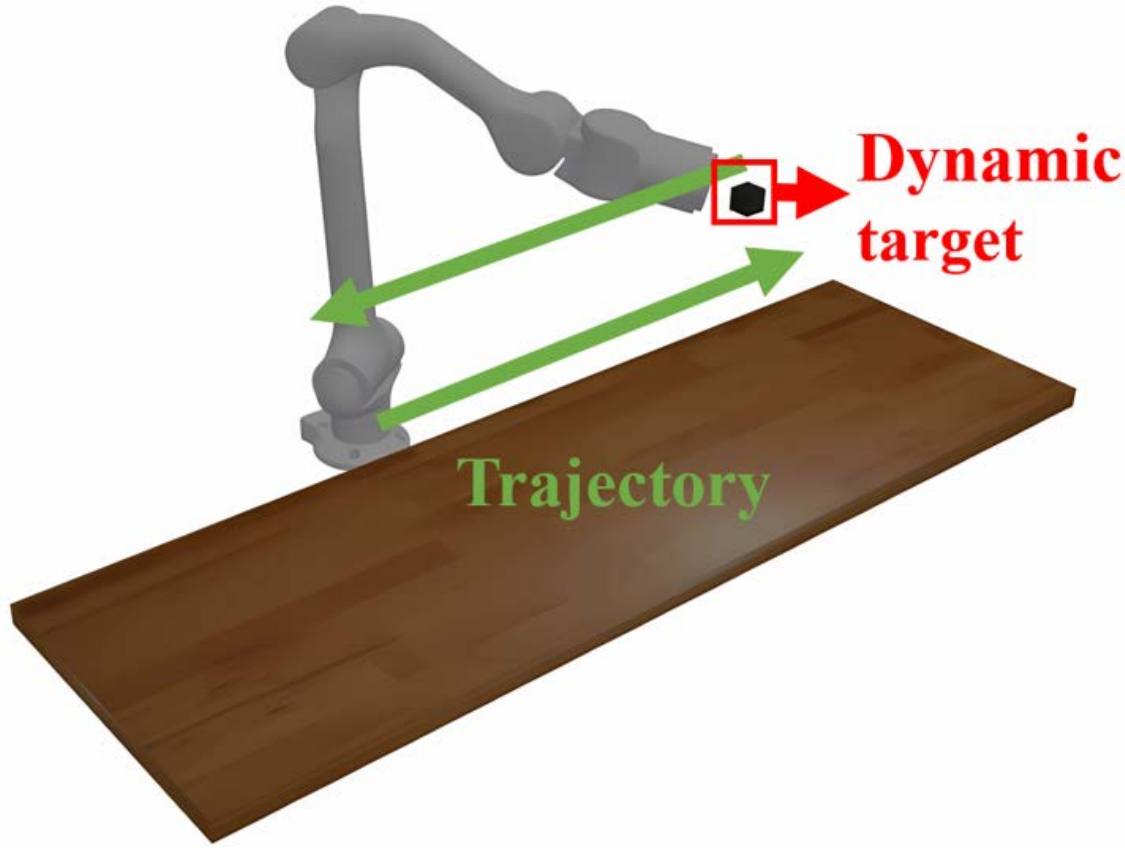}\label{dyn_tar_no_obs_setup}}
    \hspace{0.001\hsize}
    \subfigure[]{\includegraphics[width = 0.45\hsize]{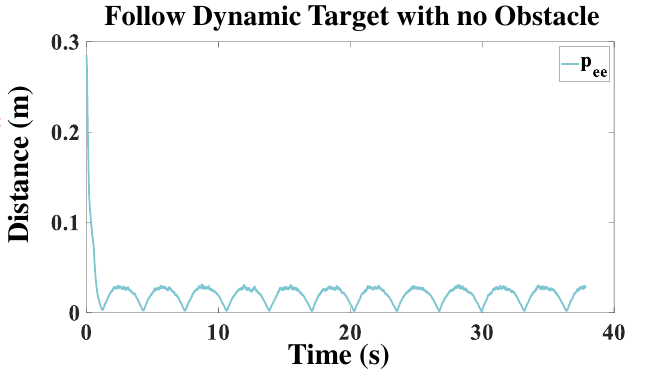}\label{dyn_tar_no_obs_result}}
    \vspace{0.001\hsize}
    \subfigure[]{\includegraphics[width = 0.3\hsize]{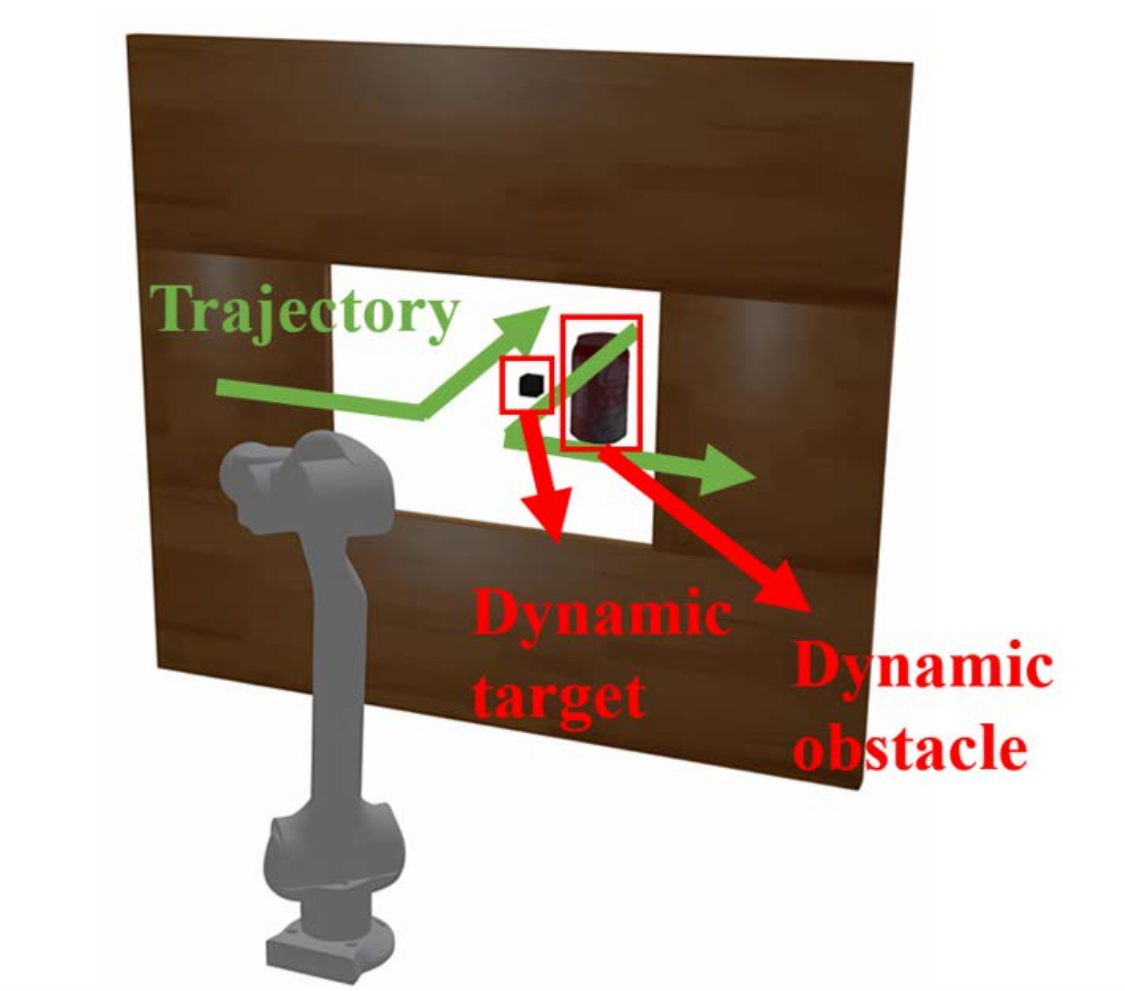}\label{dyn_tar_dyn_obs_setup}}
    \hspace{0.001\hsize}
    \subfigure[]{\includegraphics[width = 0.45\hsize]{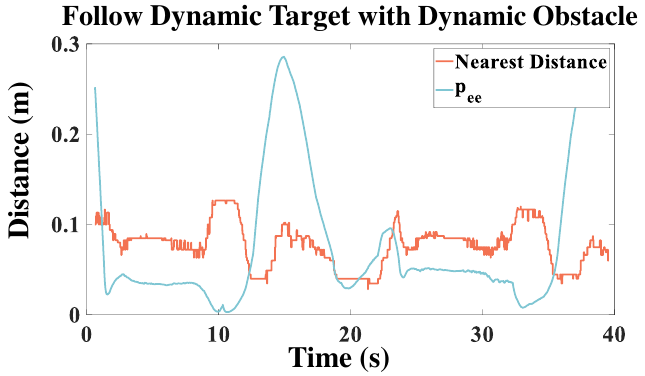}\label{dyn_tar_dyn_obs_result}}
    \caption{The simulation scene setup for the dynamic target task is shown in (a), (c), while the experimental results of the dynamic task are presented in (b), (d).  For the scenarios with obstacles, we recorded the distance of the entire manipulator from the nearest obstacle at each moment during the experiment. For the experimental scenarios with dynamic changes, we also recorded the Euclidean distance of the end of the manipulator from the target point at each moment.}
    \label{dynamic}
\end{figure*}

This experiment validates our method's ability to track dynamic target points while performing obstacle avoidance. The scenario shown in Fig. \ref{dyn_tar_no_obs_setup} tests the manipulator's effectiveness in tracking a dynamic target point within an obstacle-free environment, where the target point moves left and right at a speed of 20cm/s. Fig. \ref{dyn_tar_no_obs_result} gives the distance to the target position. The results indicate that our proposed method can track the dynamic target. Due to the upper limitation of the angular acceleration of the joints of the manipulator, an error of approximately 2cm occurs only when the target point changes direction. The scenario in Fig.~\ref{dyn_tar_dyn_obs_setup} tests the ability to track a dynamic target in an environment with both static and dynamic obstacles. Here the target point moves near the obstacle at a speed of 5cm/s, following the trajectory marked by the green line in the figure. While the dynamic obstacle moves in front of and behind the window at the same speed. Fig.~\ref{dyn_tar_dyn_obs_result} shows that throughout the experiment, the manipulator maintains a safe distance of at least 2.8cm from the obstacle and demonstrates good tracking ability when the target point is not close to the dynamic obstacle.

\begin{table}[t]
    \small\sf\centering
    \caption{The time cost of our method and the CDF method.\label{timecost_label}}
    \setlength{\tabcolsep}{4.5mm}{
        \begin{tabular}{lcc}
            \toprule
            &Mean&Root-Mean-Square\\
            \midrule
            \texttt{Ours }& 0.818ms & 1.019ms\\
            \texttt{CDF Method} &21.098ms& 21.207ms\\
        \bottomrule
    \end{tabular}
    }
\end{table}

In the real environment, we attach the marker points to a subject's dual-arms, treating each arm as a rigid body obstacle to build its SDF in advance (see Fig.~\ref{real_dy_setup}). The position and orientation of the arms are continuously acquired using a motion capture system and sent to our algorithm at a frequency of 2kHz to enable the perception of dynamic obstacles. During the experiment (see Fig.~\ref{real_dy}), the manipulator moves back and forth between two fixed target points. In the experiment, the subject is required to perform two types of actions to influence the movement of the manipulator. In the first type, the subject intentionally approached the manipulator with both arms. In response to this intentional obstruction from all directions, the manipulator could swiftly generate a real-time obstacle avoidance reflex action. In the second type, the subject's arm moves swiftly from outside the workspace to the middle of the two given targets, which obstruct the original movement of the manipulator. While our algorithm can react to the dynamic obstacles in real-time and generate actions to take another path towards the targets. Under this influence, the manipulator responds swiftly, producing motions that avoid obstacles while continuing to move toward the target point. Multimedia attachment shows the full video of the experiment.

\begin{figure*}[ht]
    \centering
    \subfigure[]{\includegraphics[width = 0.8\hsize]{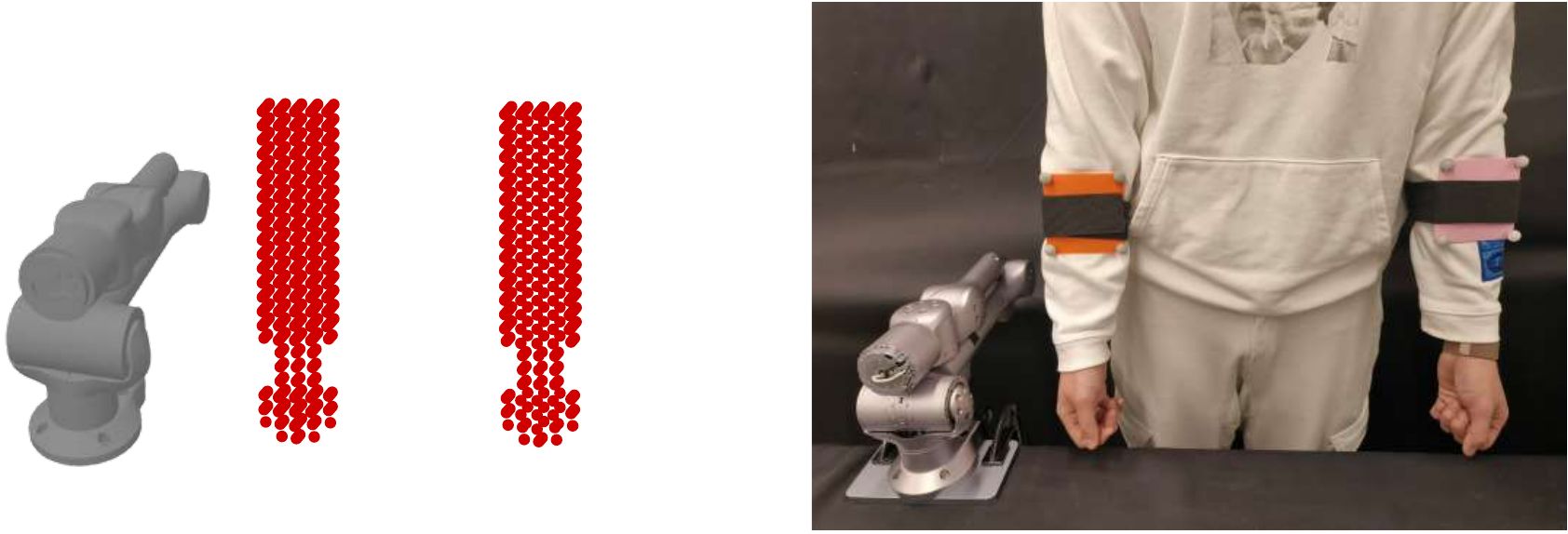}\label{real_dy_setup}}
    \subfigure[]{\includegraphics[width = 0.95\hsize]{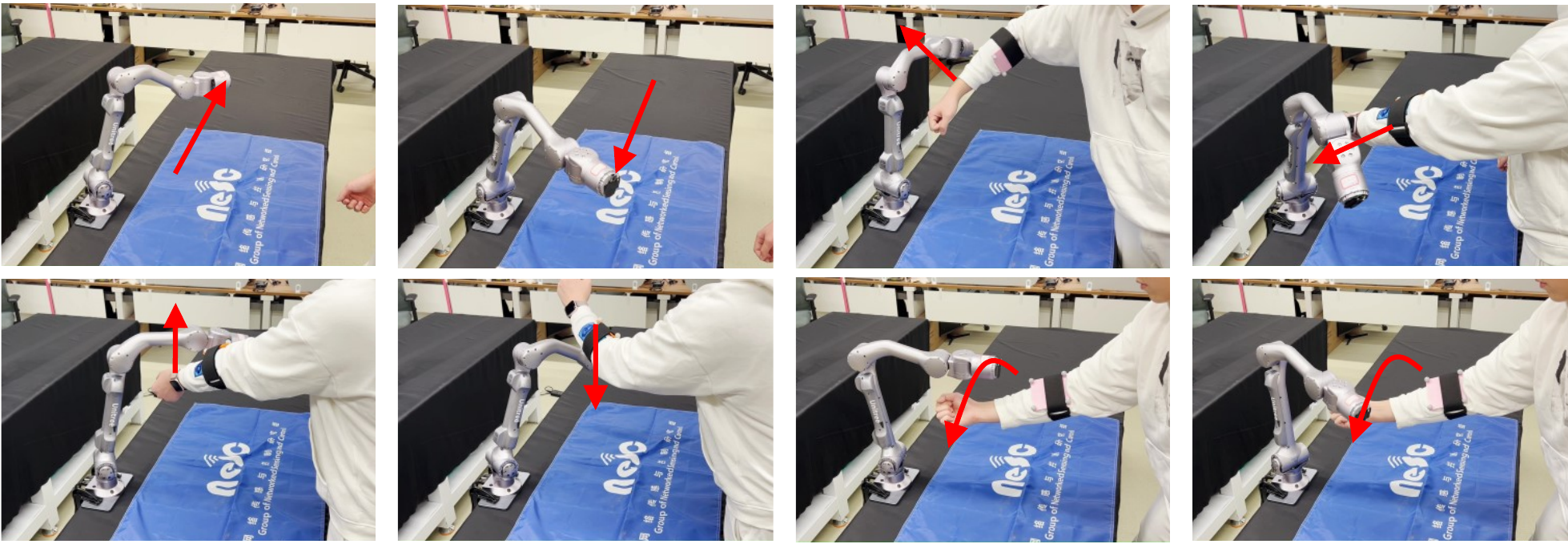}\label{real_dy}}
    \caption{We pre-computed and stored the SDF describing the geometric occupancy information of the human arms, and attach marker points from the motion capture system to the arms to obtain their real-time poses. In the experiment, the manipulator was programmed to move back and forth between two points. During this process, a human deliberately and swiftly used both arms to obstruct the movement of the manipulator. The manipulator could generate real-time obstacle avoidance reflex behaviors and successfully continued its movement toward the target point. For more experiments please refer to the supplemental video. }
    \label{real}
\end{figure*}

\section{7. Discussion }
\label{Section7_Discussion}

In this paper, we proposed an implicit framework to describes the geometric information of the manipulator and its working environment using local SDFs. Based on this framwork, we designed an online unconditioned reflex mechanism to generate the escape velocity in real-time and take actions to avoid obstacles immediately, while towards the target simultaneously. Our method offers the following advantages:

Our proposed implicit framework pre-computes and stores the geometric information of the manipulator and its working environment by 3 groups of local SDFs during the offline phase. Compared to the sampling-based algorithms, our method transforms complex distance calculations into fast table lookup operations, trading some storage space for significant improvements in time efficiency, which is clearly shown in Fig.~\ref{test8_result} and Fig.~\ref{test10_result}. Additionally, the elapsed time of our method is rarely affected with the increasing of the sizes of the working environments or the DoFs of the manipulators, because the time taken for once hash-table lookup is constant (i.e., the time complexity is $O\left(1\right)$). Although the pre-computations of these local SDFs is time-consuming, they only need to be computed once at the offline stage. Specifically, the local SDFs for the same manipulator only need to be constructed once and can be used permanently for its entire life cycle. Therefore, we believe that these local SDFs can be regarded as product attributes of the manipulator and should be provided by manufacturer.

Compared with existing SDF-based SOTA approaches, our proposed method also exhibits many benefits. Existing SDF-based planning methods either update a global SDF in every iteration or store the SDFs in a neural network. But the update for the global SDF is computationally expensive and the neural network-based method cannot adapt to scenarios that the scene changes frequently. However, in our implicit framework, we categorize the objects into 4 different levels, i.e., the permanent objects, the near-permanent objects, the manipulator, and the dynamic objects. While we use 3 different local groups of SDFs to represent the static objects and the manipulator. Through categorization and individual modeling of obstacles, the local SDFs enable the adoption of distinct strategies for different types of obstacles. For example, the SDFs for the near-permanent objects can be selectively loaded or removed according to the scene changes. In addition, the resolutions to discretize the space can be different to each other for different objects, according to needs. This design make our framework be highly flexible and extendable, and offering more potential for development in real-world applications, as well as for higher-level decision-making and other related planners.

Our proposed online unconditioned reflex mechanism effectively utilizes the separate local SDFs designed for each link of the manipulator. By mapping the desired escape velocity of each link to the null space of the manipulator, our proposed mechanism achieve simultaneous obstacle avoidance and convergence to the target. Furthermore, our mechanism does not rely on the completion of global planning; it can immediately generate actions in response to dynamic obstacles, equipping the manipulator with reflex-like responsiveness similar to human unconditioned reflexes.

Compared with existing methods, our another important contribution is the construction of $SDF_{EE}\left({\vec p}\right)$ and $SDF_{EE_i}\left({\vec p}\right)$, which are used to describe the reachability of the manipulator. Based on these SDFs, many points that has no ``collision-risk" with the manipulator can be pruned to further accelerate the online unconditioned reflex mechanism. Additionally, since our SDF could provide the distance information of any point in space from the object boundary, we can autonomously define the boundary of the reachable space of $\mathcal{L}_i$. This enhances the capability of our planning algorithm to avoid dynamic obstacles.

Although the shapes of dynamic obstacles used in the experimental tests are pre-known, our method donot limited to pre-known obstacles, since we only use the occupancy information of the obstacles. For unpredictable dynamic obstacles, $\mathcal{D}\left(t\right)$ can be perceived in real time by sensors like RGB-D camera or Lidar. In reference to paper (\cite{2024_appleARMOR_arXiv}), a large number of small laser radars are distributed and mounted on the robot's body, enabling it to comprehensively perceive the surrounding environment. This sensing system is particularly well-suited for detecting unpredictable dynamic obstacles in real-time.

The followings are the key improvements for the future work to extend our current version:

Our algorithms do not consider the acceleration/jerk constraints in its current version. However, based on Eq. \eqref{Jaco_with_constraint}, we can incorporate these constraints as an additional subtask into the Jacobian pseudo-inverse-based planning algorithm (\cite{2011_IJRR_TSR}), thereby enhancing the robustness of our planner.

Another potential improvement involves the developing of a related sensor systems. The focus of this paper is primarily on proposing a framework for describing both the environment and manipulator, as well as a planning algorithm that enables real-time obstacle avoidance action. The local SDFs used to describe the geometric information of various objects in this work are derived from their CAD models or the description files, e.g., URDF/SRDF. In the future work, a possible direction is to develop an algorithm that can autonomously construct the local SDFs from the scanning data of a RGB-D camera or a Lidar.

\section{8. Conclusion}
\label{Section8_Conclusion}

We presented an implicit framework with a real-time online unconditioned reflex mechanism which enable the manipulator to achieve a more nuanced and comprehensive understanding of their environment, while equipping them with reflex-like responsiveness similar to human unconditioned reflexes. Within this framework, we proposed three groups of local SDFs to describe the geometric information of the environments, the geometric information of each link, and the reachability of each link's end of the manipulator. These local SDFs can all be pre-computed and stored during the offline stage, enabling the real-time retrieval of escape velocity for each link. Based on the desired escape velocity, we proposed a modified geometric Jacobian matrix and then used the Jacobian-pseudo-inverse method iteration to generate real-time reflex behaviors at the online stage.

To evaluate the performance of our proposed framework and motion planning algorithm, we conducted two types of experiments. In the static environment, we compared our method with the mainstream and state-of-the-art sampling-based algorithms and achieved superior performance in terms of path planning success rate, average planning time consumption, and path quality. In dynamic environments, we set the dynamic obstacles to be avoided in real time and a dynamic target to be tracked. We recorded the nearest distance between the obstacles and the whole manipulator body, as well as the real-time position error between the dynamic target and manipulator's end effector, to assess the dynamic obstacle avoidance performance of our method.

Meanwhile, we also implemented another SDF-based SOTA approach, the CDF approach in the test scenario for comparison. The experiments demonstrated that our proposed method achieves superior obstacle avoidance performance and faster response times in complex dynamic environments using the same computing device. Our proposed method requires an average of 0.818ms to produce a single-step action, while the CDF method consumes an average of 21.098ms, which is more than 25 times slower than our method. This significant difference highlights the superior adaptability of our algorithm to dynamic environments and its closer resemblance to human unconditioned reflexes.

\bibliographystyle{SageH}
\bibliography{main}

\end{document}